\documentclass{article}
\usepackage[numbers]{natbib}
\usepackage[preprint]{neurips_2024}

\usepackage{amsmath, amssymb, mathtools}
\usepackage{amsthm}
\usepackage{xcolor, color}
\usepackage[dvipsnames]{xcolor}
\usepackage{subcaption}
\usepackage{overpic}
\usepackage{svg}
\usepackage{url}
\usepackage{float}
\usepackage{longtable}
\usepackage{hyperref}
\usepackage{tabularx}
\usepackage{csquotes}
\usepackage[linewidth=1pt]{mdframed}
\usepackage{adjustbox}
\usepackage{bbm}
\usepackage{todonotes}
\usepackage{algorithm}
\usepackage{algpseudocode}
\usepackage{booktabs}
\usepackage{comment}
\usepackage{graphicx}
\usepackage{tikz}

\usetikzlibrary{matrix,arrows}
\usetikzlibrary{arrows.meta,positioning,calc}
\tikzset{
    block/.style={draw, rectangle, fill=blue!30, minimum height=6mm, minimum width=15mm},
    vec/.style={draw, rectangle, minimum height=10mm, minimum width=12mm},
    vecno/.style={minimum height=10mm, minimum width=12mm},
    circ/.style={draw, circle, minimum size=6mm, inner sep=0pt},
    arrow/.style={-{Latex}, thick}
}
\usepackage{cleveref}
\usepackage{siunitx}

\crefname{appendix}{appendix}{appendices}
\Crefname{appendix}{Appendix}{Appendices}

\newcommand{\crefshort}[1]{%
  \begingroup
    \crefname{figure}{fig.}{figs.}%
    \crefname{section}{sec.}{secs.}%
    \crefname{equation}{eq.}{eqs.}%
    \crefname{table}{tab.}{tabs.}%
    \crefname{appendix}{app.}{apps.}%
    \cref{#1}%
  \endgroup
}

\definecolor{mypurple}{HTML}{f781bf}
\definecolor{myorange}{HTML}{ff7f00}
\definecolor{myteal}{HTML}{377eb8}

\definecolor{rev1}{HTML}{D05A45}
\definecolor{rev2}{HTML}{4F64C4}
\definecolor{both}{HTML}{B64FA3}
\definecolor{sout}{HTML}{B8872D}
\definecolor{others}{HTML}{3F9560}

\DeclareMathOperator*{\argmin}{arg\,min}
\theoremstyle{remark}
\newtheorem{remark}{Remark}

\def\tsc#1{\csdef{#1}{\textsc{\lowercase{#1}}\xspace}}
\tsc{WGM}
\tsc{QE}
\tsc{EP}
\tsc{PMS}
\tsc{BEC}
\tsc{DE}

\begin{document}

\let\WriteBookmarks\relax
\def\floatpagepagefraction{1}
\def\textpagefraction{.001}

\title{The Error of Deep Operator Networks Is the Sum of Its Parts: Branch-Trunk and Mode Error Decompositions}

\author{
  Alexander Heinlein \\
  Delft Institute of Applied Mathematics \\
  Delft University of Technology \\
  Delft, 2628CD, The Netherlands \\
  \texttt{a.heinlein@tudelft.nl}
  \And
  Johannes Taraz \\
  Delft Institute of Applied Mathematics \\
  Delft University of Technology \\
  Delft, 2628CD, The Netherlands \\
  \texttt{johannes.taraz@gmail.com}
}

\maketitle

\begin{abstract}
Operator learning has the potential to strongly impact scientific computing by learning solution operators for differential equations, potentially accelerating multi-query tasks such as design optimization and uncertainty quantification by orders of magnitude. Despite proven universal approximation properties, deep operator networks (DeepONets) often exhibit limited accuracy and generalization in practice, which hinders their adoption. Understanding these limitations is therefore crucial for further advancing the approach. 
    
This work analyzes performance limitations of the classical DeepONet architecture. It is shown that the approximation error is dominated by the branch network when the internal dimension is sufficiently large, and that the learned trunk basis can often be replaced by classical basis functions without a significant impact on performance. 

To investigate this further, a modified DeepONet is constructed in which the trunk network is replaced by the left singular vectors of the training solution matrix. This modification yields several key insights. First, for examples involving the KdV and Burgers equations, a spectral bias in the branch network is observed, with coefficients of dominant, low-frequency modes learned more effectively.
Second, through the interplay of the singular-value weighting and the optimizer's neglect of small modes, the branch error is dominated by modes with large and intermediate singular values.
Third, using a shared branch network for all mode coefficients, as in the standard architecture, improves generalization of small modes compared to a stacked architecture in which coefficients are computed separately. Finally, detrimental coupling between modes in parameter space is identified, which weakens with increasing network width.
\end{abstract}

\section{Introduction}

\paragraph{Operator learning}
Many systems in science and engineering can be formulated using differential equations, and a large body of research exists on solving them efficiently and accurately.

This work focuses on approximating the time evolution operator of time-dependent partial differential equations (PDEs), which maps the initial condition to the solution at a later time.
Classical numerical methods (e.g., finite elements with time-stepping) efficiently compute solutions for individual initial conditions but offer no efficient way to approximate the entire operator, as each initial condition requires repeated integration.

Approximating solution operators is crucial for tasks such as PDE-constrained optimization~\cite{dejong2025deepoperatorneuralnetwork, sarkar2025learningcontrolpdesdifferentiable} and uncertainty quantification~\cite{Bhattacharya}.
Operator learning (OL) methods directly approximate mappings between function spaces using neural networks~\cite{deeponet, kovachki_neural_2022, raonic_convolutional_2023}, complementing classical reduced order modeling (ROM)~\cite{MOR_benner, chaturantabut}.
OL is central to scientific machine learning, combining techniques from scientific computing and machine learning~\cite{SciML}.

\paragraph{Deep operator networks (DeepONets)}
Building on theoretical results by Chen and Chen~\cite{chen}, Lu et al.~\cite{deeponet} introduced DeepONets, which have become a standard architecture for operator learning.
The DeepONet consists of two sub-networks: the trunk network outputs $N$ basis functions evaluated at coordinate $x$, and the branch network outputs coefficients for these basis functions given a sampled initial condition. The output is a linear combination of trunk basis functions with branch coefficients. The architecture is visualized in~\cref{fig:graphical_abstract}~(top left).
\begin{figure}
    \input{GA_pdfs}
    \caption{Visualizations of the standard DeepONet (top left) and this work's main contributions: the decomposition into trunk and branch error (top right) and the modified DeepONet's architecture (bottom left) which allows us to further investigate the dominant branch error via the mode loss decomposition (bottom right). The base loss shown there is the loss of the trivial prediction $b_i=0$. \label{rev:1.4}
    For visual clarity, the bottom-right plot contains no shaded bands.
    The visualizations of the architectures are adapted and reproduced from~\cite{deeponet}.}
    \label{fig:graphical_abstract}
\end{figure}
Despite theoretical guarantees, DeepONets face practical limitations including high training data demands and limited generalizability.
Various modifications have been proposed; for a comprehensive overview see~\cite{LIU2025130518}.
Some notable modifications are Kolmogorov-Arnold-DeepONets~\cite{DeepOKAN}, improved architectures~\cite{wang_improvedarch}, two-step training with trunk orthogonalization~\cite{lee_twosteptraining_2023}, and multi-fidelity training~\cite{Howard_MF, synergy, Lu_MF}.

\paragraph{Other operator learning approaches}
Other OL approaches include learning Green's functions for interpretable solutions of linear PDEs~\cite{Boulle_GL, deep_green_learning, sun_learning_2025, melchers_neural_2025}, and addressing resolution dependence through Fourier neural operators~\cite{li_fno, kovachki_neural_2022}, convolutional neural operators~\cite{raonic_convolutional_2023, bahmani2024resolutionindependentneuraloperator, bartolucci2023representationequivalentneuraloperators}, and Laplace neural operators~\cite{cao_laplace_2024}.
Additionally, there are theoretical investigations of errors~\cite{boulle_chapter_2024, Lanthaler_errorestimates_2022, lee_twosteptraining_2023, Schwab} in DeepONets and other neural
operator architectures.

A complementary line of work incorporates governing equations directly into the learning process~\cite{Karniadakis2021}: physics-informed neural networks (PINNs)~\cite{dissanayake_neural-network-based_1994, Lagaris, raissi_physics-informed_2019} target a single PDE instance, while physics-informed DeepONets~\cite{pidnon, wang_learning_solutions, mandl, wang2021longtimeintegrationparametricevolution} extend the same idea to operator learning.
\paragraph{Related work}
Our work is inspired by the theoretical work by Lanthaler et al.~\cite{Lanthaler_errorestimates_2022} analyzing DeepONet error. Moreover, in~\cref{sec:branch}, we use a modified DeepONet: the trunk network is replaced by fixed basis vectors derived from the singular value decomposition (SVD) of the training data matrix, as this basis captures most of the variance in the data. This is (up to a scaling with the singular values and a whitening of the data) very similar to the POD-DeepONet~\cite{Lu_POD} and the SVD-DeepONet~\cite{Venturi_2023}.
We use this modified DeepONet primarily as a diagnostic tool for isolating and analyzing the branch error.
Its construction relies on an SVD of the full training data matrix, which requires that the solution be sampled on a common spatial grid for all parameter configurations, as is the case when the training data are generated by a numerical solver on a fixed mesh.
Under this assumption it constitutes a viable architecture in its own right; the SVD basis vectors $\phi_k \in \mathbb{R}^n$ are, however, defined only at the grid points. See~\cref{sec:pod_don,sec:discussion} for further discussion.
In addition, the fixed basis functions studied in~\cref{sec:trunk} are similar to the spectral neural operator~\cite{SNO}, and fixed bases (for input and output) are also used in the principal component analysis (PCA) neural operator~\cite{Bhattacharya, lanthaler_PCA}.
Beyond PCA, using the bases from nonlinear methods, such as kernel-PCA, for OL has been investigated in~\cite{eivazi2024nonlinearmodelreductionoperator}.
A direct comparison between DeepONets and ROM and their combination has been studied in~\cite{Multifidelity_residual_learning}, by training a DeepONet to learn the residual of a classical ROM.
Related to our study of the DeepONet's projection error is the investigation of the basis functions learned by a DeepONet and their use for a spectral method in~\cite{williams2024physicsinformeddeeponetslearnunderstanding}.
In this work, we also investigate spectral bias~\cite{rahaman-sb}, i.e., the tendency of neural networks to learn low-frequency components faster than high-frequency ones. While previous works, such as~\cite{khodakarami-sb, kong-sb, wang-sb, zhang-sb, fedonet}, examine spectral bias in the spatial dependency, our investigation focuses on spectral bias in the branch network, which captures the input function dependency.
Additionally, by comparing the unstacked and stacked DeepONets~\cite{deeponet}
(i.e., the architectural coupling of the modes), we establish a connection to the field of multi-task learning (MTL)~\cite{ruder2017overviewmultitasklearningdeep}.
Lastly, just as we exploit the specific DeepONet structure to decompose the error and gain insight into the network's inner workings, \cite{choi2025hybridsquaresgradientdescentmethods} leverage this same structure to develop a more efficient optimization algorithm.

\paragraph{Outline}
\Cref{sec:backgroundsetup} describes the DeepONet architecture, training/testing setup, and data.
In~\cref{sec:err_decomp_trunk_branch}, we decompose the error into trunk and branch components.
\Cref{sec:trunk} investigates the trunk network by replacing learned basis functions with fixed bases, showing that branch error dominates for large $N$, as illustrated in~\cref{fig:graphical_abstract}~(top right).
In~\cref{sec:branch}, we modify the DeepONet with a fixed SVD basis, enabling mode loss decomposition. We find that intermediate modes (those with neither dominant
nor negligible singular values)
contribute significantly to branch error, as illustrated in~\cref{fig:graphical_abstract}~(bottom right), and observe spectral bias in the branch network for the KdV equation and Burgers' equation.
This reveals, on the one hand, the improved generalization through parameter sharing, and on the other hand, the detrimental coupling between modes in DeepONets trained with gradient descent.
Furthermore, Adam mitigates gradient imbalance across singular value scales.
\Cref{sec:discussion} concludes with discussion and limitations. \Cref{tab:notation} contains notation.
To summarize, our three main contributions are:
\begin{enumerate}
    \item We systematically study trunk and branch errors with respect to the inner dimension $N$ for practical examples.
    \item We decompose the branch error into modes, showing significant contributions of the large and intermediate modes.
    \item The decomposition enables analysis of spectral bias in the branch network and mode coupling effects.
\end{enumerate}

The code and data from this study are available at \url{https://github.com/jotaraz/ModeDecomposition-DeepONets}.

\section{Glossary}

\begin{longtable}{rp{0.55\textwidth}p{0.25\textwidth}p{0.05\textwidth}}
    \label{tab:notation}                                                                                                                                                                                                                          \\
    \toprule
    \textbf{symbol}            & \textbf{name}                                                                                                               & \textbf{note}                                                                      \\  \midrule
    $Z^+$                      & Moore--Penrose inverse of matrix $Z$                                                                                        &                                                                                    \\
    $\text{ran}(Z)$            & range / column space of matrix $Z$                                                                                          &                                                                                    \\
    $||Z||_F$                  & Frobenius norm                                                                                                              & $||Z||_F \coloneqq \sqrt{\sum_{ij} Z_{ij}^2}$                                      \\
    $N$                        & number of output neurons of branch and trunk net (``inner dimension'')                                                      & $N \in \mathbb{N}$ & \crefshort{eq:deeponet}                                                                 \\
    $\mathcal{X}$              & spatial domain of interest                                                                                                  & $\mathcal{X} \subset \mathbb{R}^d$ & \crefshort{sec:pde_setup}                                                 \\
    $\mathcal{C}(\mathcal{X})$ & set of continuous functions on $\mathcal{X}$                                                                                &                                                                                    \\
    $u$                        & solution function of time-dependent PDE                                                                                     & $u \in \mathcal{C}(\mathcal{X} \times [0, \tau])$ & \crefshort{sec:pde_setup}                                  \\
    $p$                        & input function (e.g., initial condition of time-dependent PDE), later also used for discretized input function              & $p(\cdot) = u(\cdot, t=0) \in \mathcal{C}(\mathcal{X})$ & \crefshort{sec:pde_setup}                            \\
    $\tau$ & evolution time & $p(\cdot) \mapsto u(\cdot, \tau)$ & \crefshort{sec:pde_setup} \\
    $m$                        & number of input functions in a data set                                                                                     & $m \in \mathbb{N}$ & \crefshort{sec:framework}                                                                \\
    $M$                        & number of sampling points of $p$                                                                                            & $M \in \mathbb{N}$ & \crefshort{sec:deeponet}                                                          \\
    $\Bar{x}_j$                & $j$-th sampling point of $p$ with $1 \leq j \leq M$                                                                         & $\Bar{x}_j \in \mathcal{X}$ & \crefshort{sec:deeponet}                                                          \\
    $\hat{p}$                  & discretized input function (sampled at $\Bar{x}_j$), later $p$ is used to denote this too                                         & $\hat{p} \in \mathbb{R}^M$ & \crefshort{sec:deeponet}                                                           \\
    $G$                        & true solution operator                                                                                                      & $G : p(\cdot) \mapsto u(\cdot, t=\tau)$ & \crefshort{sec:pde_setup}                                             \\
    $\tilde{G}_\theta$         & approximation of $G$ (using a DeepONet)                                                                                     & & \crefshort{sec:pde_setup}                                                                                    \\
    $x$                        & coordinate at which $G(p)(\cdot)$ and $\tilde{G}_\theta(\hat{p})(\cdot)$ are evaluated                                      & $x \in \mathcal{X}$                                                                \\
    $n$                        & number of coordinates to evaluate $G(p)(\cdot)$ and $\tilde{G}_\theta(\hat{p})(\cdot)$ in test and training data            & $n \in \mathbb{N}$ & \crefshort{sec:framework}                                                                  \\
    $A$                        & target data matrix                                                                                                          & $A \in \mathbb{R}^{n \times m}$ & \crefshort{sec:framework}                                                     \\
    $t_k$                      & $k$-th trunk output neuron: either a function $t_k(x)$ or a vector $t_k~=~(t_k(x_1)~\ldots~t_k(x_n))^T$                     & $t_k \in \mathcal{C}(\mathcal{X})$ or $t_k \in \mathbb{R}^n$ & \crefshort{sec:deeponet}                       \\
    $T$                        & trunk output matrix                                                                                                         & $T_{ik} = t_k(x_i), T \in \mathbb{R}^{n \times N}$ & \crefshort{sec:framework}                                 \\
    $b_k$                      & $k$-th branch output neuron: either a function $b_k(p)$ or a vector $b_k~=~(b_k(p_1)~\ldots~b_k(p_m))^T$                    & $b_k \in \mathcal{C}\left(\mathbb{R}^M\right)$ or $b_k \in \mathbb{R}^m$ & \crefshort{sec:deeponet} \\
    $B$                        & branch output matrix                                                                                                        & $B_{jk} = b_k(p_j), B \in \mathbb{R}^{m \times N}$ & \crefshort{sec:framework}                                 \\
    $\tilde{A}$                & approximation of $A$ by the DeepONet                                                                                        & $\tilde{A} = TB^T \in \mathbb{R}^{n \times m}$ & \crefshort{sec:framework}                                     \\
    $\varepsilon$              & squared total error of DeepONet approximation                                                                               & $\varepsilon \coloneqq ||A-\tilde{A}||_F^2$ & \crefshort{sec:trunk_and_branch_errors}                                        \\
    $\delta$                   & relative error of DeepONet approximation                                                                                    & $\delta := \sqrt{\varepsilon} / ||A||_F$                                                     \\
    $X_{tr}$                   & training variant of a variable $X$ (e.g., $X=\varepsilon, A, ...$)                                                          &                                                                                    \\
    $X_{te}$                   & test variant of a variable $X$                                                                                              &                                                                                    \\
    $\sigma_k$                 & $k$-th singular value of $A_{tr}$ (ordered descendingly)                                                                    & $\Sigma = \text{diag}(\sigma_1, \ldots, \sigma_r)$ & \crefshort{eq:Asvd}                                                                                   \\
    $\phi_k$                   & $k$-th mode / $k$-th left singular vector of $A_{tr}$                                                                       & $\phi_k \in \mathbb{R}^n, \Phi = [\phi_1~\ldots~\phi_r]$ & \crefshort{eq:Asvd}                                                          \\
    $v_k$                      & optimal coefficients for the $k$-th mode and the training data's input functions / $k$-th right singular vector of $A_{tr}$ & $v_k \in \mathbb{R}^{m_{tr}}, V = [v_1~\ldots~v_r]$ & \crefshort{eq:Asvd}                                                      \\
    $w_k$                      & optimal coefficients for the $k$-th mode and the test data's input functions                                                & $w_k \in \mathbb{R}^{m_{te}}$                                                      \\
    $\mathcal{L}$              & loss / mean-squared error of DeepONet approximation                                                                         & $\varepsilon/(nm)$                                                                \\
    $L_{k,tr}$                 & unweighted training loss of mode $k$                                                                                        & $||b_{k,tr}-v_k||_2^2$ & \crefshort{eq:modelossdecompose_train}                                                             \\
    $L_{k,te}$                 & unweighted test loss of mode $k$ (normalized for comparability to $L_{k,tr}$)                                               & $\frac{m_{tr}}{m_{te}} ||b_{k,te} - w_k||_2^2$ & \crefshort{eq:Li_test}                                    \\
    $\sigma_k^2 L_k$           & weighted loss of mode $k$                                                                                                   & & \crefshort{sec:modelossdecomp-deriv}                                                                                  \\
                    --           & unweighted base loss / mode loss $L_k$ for $b_k = 0$                                                                        & $1$ (train), $\frac{m_{tr}}{m_{te}} ||w_k||_2^2$ (test) & \crefshort{sec:modelossdecomp-deriv}                             \\
                 --              & weighted base loss / mode loss $ \sigma_k^2 L_k$ for $b_k = 0$                                                              & $\sigma_k^2$ (train), $\sigma_k^2 \frac{m_{tr}}{m_{te}} ||w_k||_2^2$ (test) & \crefshort{sec:modelossdecomp-deriv}        \\
    \bottomrule
    \caption{Notation frequently used in this work.}
\end{longtable}

\section{Background and Setup}
\label{sec:backgroundsetup}

This section provides background on neural networks, the problem setup (time evolution under a PDE), the DeepONet architecture, and the training/testing framework.

\subsection{Notation for Neural Networks and Their Training}
\label{sec:nns}

We fix notation for neural networks and the algorithms used to train them; for a
comprehensive introduction, see~\cite{deeplearning}.

Consider approximating a function $f$ by a parametric model $\tilde{f}_\theta$ with
parameters $\theta$. Given inputs $\{x_1, \ldots, x_m\}$ and targets $y_i = f(x_i)$,
training minimizes the mean-squared error (training loss)
\begin{align*}
    \mathcal{L}_{tr}(\theta) = \frac{1}{m} \sum_{i=1}^m |y_i - \tilde{f}_\theta(x_i)|^2 .
\end{align*}
As models we use multi-layer perceptrons (MLPs), which are universal approximators of
continuous functions~\cite{cybenko, pinkus_mlp_approx}. An MLP alternates affine maps
with a componentwise non-polynomial activation function $\sigma$; the vector $\theta$
collects all its weights and biases. Each component of a hidden layer is called a
        \textit{neuron}; the number of neurons in a layer is its \textit{width}, and the number
of hidden layers $D$ is the \textit{depth}. All hidden layers in this work share the
same width $w$. We use the GELU activation~\cite{hendrycks2023gaussianerrorlinearunits};
see~\cref{app:hyperparams} for its definition and for our reasons for choosing it.

Gradient descent (GD) updates the parameters iteratively,
\begin{align} \label{eq:gd}
    \theta_{t+1} = \theta_t - \alpha_t \nabla \mathcal{L}_{tr}(\theta_t),
\end{align}
with learning rate $\alpha_t$~\cite{init1, init2}, which may follow a schedule.
        Unless stated otherwise, we use Adam~\cite{adam} instead, which combines
momentum~\cite{Rumelhart1986LearningIR} with adaptive, per-parameter learning
rates~\cite{adagrad, RMSprop}; its update rule and hyperparameters are given
in~\cref{app:hyperparams}.

\subsection{General Problem Setup}
\label{sec:pde_setup}

Consider the time-dependent PDE on spatial domain $\mathcal{X} \subset \mathbb{R}^d$ and time interval $(0, \tau]$:
\begin{alignat*}{2}
    \frac{\partial u}{\partial t} &= \mathcal{D}(u), &\qquad& \text{in } \mathcal{X} \times (0, \tau],         \\
    \mathcal{B}(u)                &= 0,              &\qquad& \text{on } \partial\mathcal{X} \times (0, \tau], \\
    u(\cdot, 0)                   &= p(\cdot),       &\qquad& \text{in } \mathcal{X},
\end{alignat*}
where $\mathcal{D}$ is a (possibly nonlinear) differential operator, $\mathcal{B}$ denotes a suitable boundary condition on $\partial\mathcal{X}$, and $p \in \mathcal{C}(\mathcal{X})$ is the initial condition.
Under appropriate regularity assumptions on $\mathcal{X}$, $\mathcal{D}$, $\mathcal{B}$, and $p$, this PDE admits a unique solution $u \in \mathcal{C}(\mathcal{X} \times [0, \tau])$.
We then define the time evolution (or solution) operator
\begin{align*}
    G : \mathcal{C}(\mathcal{X}) & \to \mathcal{C}(\mathcal{X}), \quad p(\cdot) \mapsto u(\cdot, \tau),
\end{align*}
which maps the initial condition to the solution at time $\tau$.

\begin{remark}
The function $p$ plays multiple roles: it is the initial condition of the PDE, the input to the solution operator $G$, and the function that parametrizes the solution. Correspondingly, the operator learning literature uses the terms initial condition, input function, and parameter function interchangeably. We adopt input function throughout this work.
\end{remark}

\subsection{DeepONet}
\label{sec:deeponet}
The DeepONet, as introduced in \cite{deeponet}, is designed to approximate such solution operators.
It can be seen as a deep neural network realization of the architecture in the generalized universal approximation theorem for operators in~\cite{chen}.
To make the infinite dimensional problem, that is, mapping $p(\cdot) \in \mathcal{C}(\mathcal{X})$ to $u(\cdot, \tau) \in \mathcal{C}(\mathcal{X})$, computationally accessible, the initial condition $p$ is encoded in a finite dimensional vector space as $\hat{p} = \left(p(\bar{x}_1)~\ldots~p(\bar{x}_M) \right)^T \in \mathbb{R}^M$ by sampling at $M$ locations.
This potentially involves a loss of information, which we will also address in~\cref{sec:framework}.
As mentioned briefly before, the DeepONet is a neural network consisting of two sub-networks: the trunk network outputs $t_k(x)$ given coordinate $x$, and the branch network outputs $b_k(\hat{p})$ given $\hat{p}$. The final output is
\begin{align} \label{eq:deeponet}
    \tilde{G}_\theta(\hat{p})(x)  = \sum_{k=1}^N b_k(\hat{p}) t_k(x),
\end{align}
where $\theta$ are parameters and $N$ is the number of output neurons of both trunk and branch network; see also~\cref{fig:graphical_abstract} (top left).
Equivalently, the DeepONet's output is a linear combination of the $N$ trunk output neurons $t_k(x)$ with the branch network's output neurons $b_k(\hat{p})$ as coefficients.
Thus, all DeepONet output functions $\tilde{G}_\theta(\hat{p})(\cdot)$ lie in the \textit{trunk space} $\mathcal{T}(N) \coloneqq \text{span}\{ t_k \}_{k=1}^N \subset \mathcal{C}(\mathcal{X})$.
The dimension of $\mathcal{T}(N)$ is at most $N$, with equality when the trunk functions $\{ t_k \}_{k=1}^N$ are linearly independent; in practice, this is typically the case.
For simplicity, we call $N$ the \textit{inner dimension of the DeepONet} and we denote the trunk functions as basis functions.

Both trunk and branch network are usually implemented as MLPs, but other choices are possible, as previously mentioned.
We focus on this standard architecture with MLP-based designs; analyzing more advanced DeepONet variants is left to future work. Our experiments use a fixed hidden-layer width $w$ throughout, but this is a convenience rather than a necessity: our error decomposition (Sections~\ref{sec:err_decomp_trunk_branch} and~\ref{sec:branch}) depends only on the inner dimension $N$ and so carries over unchanged to varying-width (e.g., funnel or bottleneck) architectures, though these may quantitatively affect the empirical mode-error distribution.
The parameters of both sub-networks together form $\theta$, the DeepONet's parameters.
These are found by minimizing the mean-squared loss
\begin{align*}
    \mathcal{L}(\theta)  = \frac{1}{K} \sum_{i=1}^K |G(p_i)(x_i) - \tilde{G}_\theta(\hat{p}_i)(x_i)|^2,
\end{align*}
over $K$ samples of input functions $p_i$ and evaluation coordinates $x_i$.

\subsection{Setup for Training and Testing}
\label{sec:framework}

Our setup uses:
\begin{enumerate}
    \item Initial conditions from a finite-dimensional subspace with sufficient sampling ($M$ large) such that $p$ is reconstructible from $\hat{p}$. For simplicity, we thus stop distinguishing between $p$ and $\hat{p}$ (details in~\cref{app:initcondencoding}).
    \item Discretization with $m$ initial conditions $\left(p_j\right)_{j=1,\ldots,m}$ and $n$ evaluation coordinates $\left(x_i\right)_{i=1,\ldots,n}$, yielding $K = nm$ training points, i.e., the full tensor product of these discretizations.
    \item Same evaluation coordinates for training and testing ($n_{tr} = n_{te} = n$), focusing on generalization in parameter space. We study the more general case in~\cref{app:offgrid}, where we find similar errors on- and off-grid.
\end{enumerate}

We arrange the targets (or true solutions) in a matrix $A_{ij} = G(p_j)(x_i)$, such that $A \in \mathbb{R}^{n \times m}$. The DeepONet output is
\begin{align*}
    \tilde{A}_{ij} = \tilde{G}_\theta(p_j)(x_i) = \sum_{k=1}^N t_k(x_i) b_k(p_j) \Longleftrightarrow
    \tilde{A} = TB^T,
\end{align*}
where $T_{ik} = t_k(x_i) \in \mathbb{R}^{n \times N}$ and $B_{jk} = b_k(p_j) \in \mathbb{R}^{m \times N}$ are trunk and branch outputs. The loss is $\mathcal{L}(\theta)  = \frac{1}{K} ||A-\tilde{A}||_F^2$ with Frobenius norm $||Z||_F^2 = \sum_{ij} z_{ij}^2$.
This matrix notation simplifies analysis and enables efficient computation by evaluating $T$ once for all inputs.

\subsection{Data and Experiments}
\label{sec:data}

In this work, we train DeepONets to approximate the time evolution operators that map the initial condition to the solution at a given final time $\tau$. Our standard example problem is the KdV equation, similar to~\cite{williams2024physicsinformeddeeponetslearnunderstanding},
\begin{alignat*}{9}
0                                       &= \frac{\partial u}{\partial t} + \frac{u}{2 \pi} \frac{\partial u}{\partial x} +  \frac{0.01}{8 \pi^3} \frac{\partial^3 u}{\partial x^3}
& & \quad \forall x \in (0, 1), t > 0, & \quad &&
\frac{\partial^2 u}{\partial x^2}(0, t) &= \frac{\partial^2 u}{\partial x^2}(1, t)
& & \quad \forall t \geq 0,       \\
u(0, t)                                 &= u(1, t)
& & \quad \forall t \geq 0, & \quad &&
u(x, t=0)                               &= p(x)
&  & \quad \forall x \in (0, 1), \\
\frac{\partial u}{\partial x}(0, t)     &= \frac{\partial u}{\partial x}(1, t)
& & \quad \forall t \geq 0, & \quad &&
p(x)                                    &= \sum_{k=1}^{5} a_k \sin(2 \pi k x)
& & \quad \forall x \in (0, 1),
\end{alignat*}

with $\tau = 0.2$.
Most results shown in this work are computed for this standard example.
We also consider the advection-diffusion equation and Burgers' equation, training separate operators for each PDE and value of $\tau$.
Unless stated otherwise, the observations reported apply qualitatively to all example problems described in~\cref{app:data}; in general, we report results only for the KdV equation and include additional examples when they offer new insights or qualitative differences. Unless indicated otherwise, all figures show averages over ten DeepONets with different initializations. Where this does not impair visual clarity, we also include shaded bands spanning $\pm1$ standard deviation across these DeepONets. In general, mode-loss distributions, such as those in~\cref{fig:modeerrors_GD,fig:modeerrors_Adam,fig:stacked_modelosses}, do not include shaded bands; \cref{fig:spectralbias_branch} is an exception.
        If the training and test curves are too close to show both bands, we include only the test band.

\section{Error Decomposition Into Trunk and Branch Errors}
\label{sec:err_decomp_trunk_branch}

In this section, we define the projection error and formalize how the total approximation error is distributed between trunk and branch.
As mentioned in~\cref{sec:framework}, we consider a fixed evaluation grid for the trunk network as this simplifies the further discussion.
Thus, we now use the terms basis functions and basis vectors (obtained through the evaluation of the basis functions on the grid) interchangeably.

\subsection{Projection Error}

Consider a set of vectors $\{ a_i \}_{i=1}^m \subset \mathbb{R}^n$ and a basis $\{ t_k \}_{k=1}^N \subset \mathbb{R}^n$.
The error of projecting $\{ a_i \}_{i=1}^m$ onto the space spanned by $\{ t_k \}_{k=1}^N \subset \mathbb{R}^n$, or the \textit{projection error}, quantifies how well the $a_i$ can be reconstructed through optimal linear combinations of the $t_k$.
For accurate reconstruction, this error should be small.
More formally,  we define the projection error as
\begin{align*}
    \varepsilon_P(T,A) & := ||(I-TT^+)A||_F^2,
\end{align*}
where $T=[t_1~\ldots~t_N] \in \mathbb{R}^{n \times N}$, $A = [a_1~\ldots~a_m] \in \mathbb{R}^{n \times m}$, and $T^+$ denotes the Moore-Penrose inverse of $T$. Here, $TT^+$ represents the orthogonal projection onto the span of the $\{ t_k \}_{k=1}^N$. Thus, $(I-TT^+)A$ captures the part of $A$ that cannot be represented by the $t_k$~\cite{golub}.

\subsection{Trunk and Branch Errors}
\label{sec:trunk_and_branch_errors}

We adapt the error decomposition framework of Lanthaler et al.~\cite{Lanthaler_errorestimates_2022} with two simplifications: we eliminate the encoding error by ensuring $p$ can be reconstructed from $\hat{p}$, as described in~\cref{sec:framework}, and we define the error over a finite sample set rather than a probability measure over the input space.
This allows for a more direct analysis of how the total approximation error splits between trunk and branch networks.
We consider the difference between the DeepONet's output $\tilde{A}$ and the target data matrix $A$:
\begin{align*}
    \tilde{A} - A = T B^T - A = T B^T - T T^+ A + T T^+ A - A = T (B^T - T^+ A) + (T T^+ -I) A.
\end{align*}
Since $T (B^T - T^+ A)$ and $(T T^+ -I) A$ are orthogonal to each other, this yields the error decomposition
\begin{align}
    \varepsilon := ||\tilde{A} - A||_F^2 & = \underbrace{||T (B^T - T^+ A)||_F^2}_{=: \varepsilon_B} + \underbrace{||(T T^+ -I) A||_F^2}_{=: \varepsilon_T = \varepsilon_P(T,A)}. \label{eq:myerrordecomp}
\end{align}
In the following sections, we investigate the different error components $\varepsilon_B$ and $\varepsilon_T$.

\subsubsection{Trunk Error}
\label{sec:trunktheory}
\Cref{eq:myerrordecomp} thus defines the trunk error $\varepsilon_T$ as
the projection error $\varepsilon_P(T,A)$ with the trunk matrix $T$, that is,
the error of projecting the target data matrix onto the trunk space, cf.~\cref{sec:deeponet}.
This formally proves that a small projection error is necessary for a small error $\varepsilon$, but in general, not sufficient.
Now, we consider the SVD of the training data matrix
\begin{align}
    A & = \Phi \Sigma V^T = [\Phi_1~\Phi_2]
    \begin{bmatrix}
        \Sigma_1 & 0 \\ 0 & \Sigma_2
    \end{bmatrix}
    \begin{bmatrix}
        V_1^T \\ V_2^T
    \end{bmatrix}
    = \Phi_1 \Sigma_1 V_1^T + \Phi_2 \Sigma_2 V_2^T, \label{eq:Asvd}
\end{align}
where, for the rank $r$ of $A$, the blocks
$\Phi_1 \in \mathbb{R}^{n \times N}$, $\Sigma_1 \in \mathbb{R}^{N \times N}$,
$V_1 \in \mathbb{R}^{m \times N}$ collect the $N$ largest singular values and their
left and right singular vectors, and
$\Phi_2 \in \mathbb{R}^{n \times (r-N)}$, $\Sigma_2 \in \mathbb{R}^{(r-N) \times (r-N)}$,
$V_2 \in \mathbb{R}^{m \times (r-N)}$ collect the remaining $r-N$ singular components.
Thus, $\Phi_1 \Sigma_1 V_1^T$ is the truncated SVD of $A$.
Then, the Eckart-Young-Mirsky theorem~\cite{Eckart_Young_1936} states that the projection error for all matrices $T \in \mathbb{R}^{n \times N}$ satisfies $\varepsilon_P(T,A) \geq ||\Sigma_2||_F^2$, with equality achieved by $T = \Phi_1 C$ for any full-rank matrix $C\in \mathbb{R}^{N \times N}$.
Intuitively, this is because the first $N$ left singular vectors of $A$ capture the $N$ principal directions of $A$, that is, the $N$ directions with the highest variance.

\subsubsection{Branch Error}
\label{sec:branchtheory}
To better understand the branch error $\varepsilon_B$, we first compute the branch matrix $B_*$, which is optimal in the following sense:
given a trunk matrix $T$, $B_*$ is the branch matrix for which $TB^T$ best approximates $A$ in the Frobenius norm.
\begin{align*}
    B_* & = \argmin_{B \in \mathbb{R}^{m \times N}} ||A-TB^T||_F^2 = \argmin_{B \in \mathbb{R}^{m \times N}} ||(I-TT^+)A + TT^+A - TB^T||_F^2                                         \\
        & = \argmin_{B \in \mathbb{R}^{m \times N}} \left( ||(I-TT^+)A||_F^2 + ||T(T^+A-B^T)||_F^2 \right) = \argmin_{B \in \mathbb{R}^{m \times N}} ||T(T^+A-B^T)||_F^2 = (T^+ A)^T.
\end{align*}
Thus, when investigating the branch network, one might intuitively define the branch error as
\begin{align}
    \varepsilon_C & := ||B-B_*||_F^2 = ||B-(T^+ A)^T||_F^2 = ||B^T-T^+ A||_F^2, \label{eq:epsc}
\end{align}
that is, the difference between the actual branch matrix $B$ and the optimal branch matrix $B_*$.
The inequality
\begin{align}
    \varepsilon & \leq ||T||_2^2 \underbrace{||B^T - T^+ A||_F^2}_{\varepsilon_C} + ||(T T^+ -I) A||_F^2 \label{eq:lanthaler}
\end{align}
can be derived as a bound on $\varepsilon$.
While $\varepsilon_C$ puts the same weight on the error of each branch neuron, $\varepsilon_B = ||T(B^T-T^+A)||_F^2$ accounts for the trunk matrix $T$, weighting the approximation error of each branch neuron (column of $B$) by the norm of the corresponding trunk neuron.
Hence, we use $\varepsilon_B$ as the branch error.

\begin{remark}
In practice,~\cref{eq:lanthaler} is a very loose bound, since the norms of the trunk neurons vary strongly, cf.~\cref{app:branch_error_bound}.
\end{remark}

\section{Investigating the Trunk Network}
\label{sec:trunk}

The error decomposition enables systematic investigation of the trunk network. Many trunk modifications exist: orthogonalization~\cite{lee_twosteptraining_2023}, Fourier features~\cite{Lu_POD}, SVD basis~\cite{Bhattacharya, Lu_POD}, Chebyshev or trigonometric functions~\cite{SNO}, and transfer learning~\cite{williams2024physicsinformeddeeponetslearnunderstanding}.
We investigate: (1) How do prescribed bases compare to learned bases? (2) Does the learned trunk approach optimal trunk error? (3) How is the error distributed between trunk and branch?

We compare different bases $t_k$, either learned or prescribed, but the output remains $\sum_{k=1}^N t_k(x) b_k(p)$. We test (details in~\cref{app:bases}):
Learned (standard DeepONet), SVD ($t_k$ is $k$-th left singular vector of $A$), Legendre ($k$-th Legendre polynomial), Chebyshev ($k$-th Chebyshev polynomial), Cosine ($t_k(x)=\cos((k-1)\pi x)$).
\begin{figure}[t]
    \centering
    \includegraphics{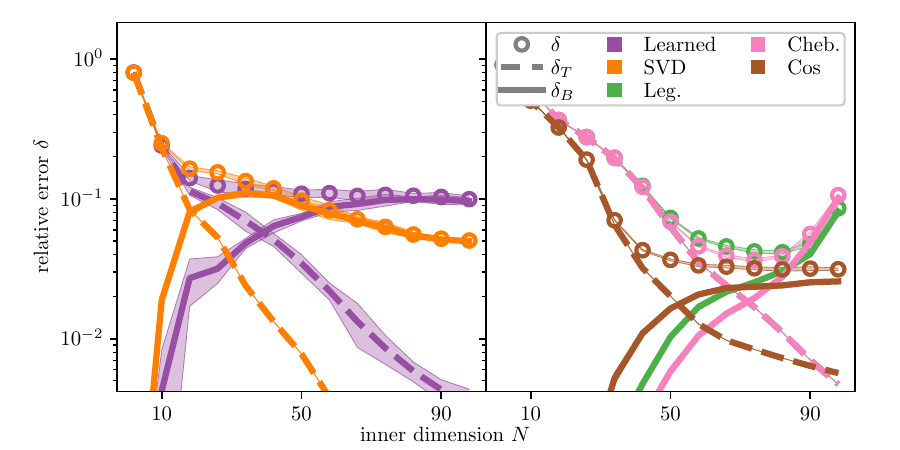}
    \caption{\textbf{Relative total, trunk and branch test errors of DeepONets with various bases plotted over $N$.}
        \textbf{Colors:} Purple lines: learned trunk (standard DeepONet), orange: SVD basis, green: Legendre polynomials, pink: Chebyshev polynomials, brown: cosine.
        \textbf{Line-symbols:} Total error $\delta$ (circle), trunk error $\delta_T$ (dashed), branch error $\delta_B$ (solid).
    The relative error is defined in the glossary,~\cref{tab:notation}. Since the Legendre and the Chebyshev polynomials span the same space, they have identical trunk errors; thus only one curve is visible.
    }
    \label{fig:differentbases}
\end{figure}

\Cref{fig:differentbases} shows errors for different bases. For small $N$, learned and SVD bases perform best. The learned basis plateaus around $N=20$, while SVD error decreases slowly.
Legendre and Chebyshev total errors decrease until $N=70$, then increase.
For large $N$, cosine achieves lowest total error. Across examples, all bases match or exceed learned basis performance for large $N$.

The SVD basis achieves the lowest trunk error on training data by the Eckart-Young-Mirsky theorem~\cite{Eckart_Young_1936}.
\Cref{fig:differentbases} shows empirically that this translates to the test data as well; see~\cref{app:errordecompose_test_derive} for further discussion. However, optimal projection error does not guarantee optimal total error; low branch error is also necessary. We find that there exists an $N$ (in our cases $N \approx 50$) such that the standard DeepONet learns trunk functions spanning the solutions well, with the trunk error becoming negligible in comparison to the dominant branch error. For the other bases, we also find dominant branch errors for $N \gtrsim 50$.
We stress that this comparison of prescribed and learned bases is intended to characterize what the learned trunk network represents, namely a basis spanning essentially the same space as the leading SVD modes, rather than to recommend prescribed bases as a replacement for the learned trunk in practice. In particular, the SVD basis serves here only to facilitate the analysis: it presupposes a full training data matrix on a common grid and is therefore not proposed as a generally deployable trunk; see~\cref{sec:pod_don,sec:discussion}.
    The increase of the branch error with $N$ can intuitively be understood; approximating ten coefficients is easier than ninety using a fixed-capacity neural network.
Since the branch error dominates, we study it in detail next.
Note that we obtain similar results for the 2D heat equation with different input functions in~\cref{app:2d_trunk_branch}.

\begin{remark}
\label{remark:choose_N_based_on_sigma}
Since the trunk error of the SVD basis is $||\Sigma_2||_F^2$, the singular values of $A_{tr}$ determine, without any training, the smallest $N$ at which a target accuracy is not already precluded by the trunk error alone.
Given the dominance of the branch error, this is necessary but far from sufficient: for $N=90$ in~\cref{tab:branch_error_bounds}, the trunk error is roughly three orders of magnitude below the total error.
\end{remark}
\section{Investigating the Branch Network}
\label{sec:branch}

In~\cref{sec:trunk}, we observed that, for our examples and large $N$, the branch error dominates (i.e., the coefficient approximation is poor). In this section, we study whether all coefficients are poorly approximated or only those for certain basis vectors, and why.
To this end, we introduce the DeepONet with the fixed SVD basis in~\cref{sec:pod_don}. In~\cref{sec:modelossdecomp-deriv}, we decompose the branch error into individual coefficient errors, which we also refer to as mode losses. Next, in~\cref{sec:modelossdistr}, we analyze the distribution of the mode losses in practice. Lastly, in~\cref{sec:why}, we study phenomena that affect the coefficient error distribution, such as spectral bias and inter-coefficient coupling.

\subsection{Modified DeepONet}
\label{sec:pod_don}

\begin{longtable}{lrcl}
    \label{tab:don_notation}                             \\
    \toprule
                    & \textbf{matrix formulation}                                         &     & \textbf{individual evaluation}              \\ \midrule
    true solution (train)  & $\left( \Phi_1 \Sigma_1 V_1^T + \Phi_2 \Sigma_2 V_2^T \right)_{ij}$ & $=$ & $\sum_{k=1}^r \sigma_k (\phi_k)_i (v_k)_j$  \\
    true solution (test)  & $\left( \Phi_1 \Sigma_1 W_1^T + \Phi_2 \Sigma_2 W_2^T \right)_{ij}$ & $=$ & $\sum_{k=1}^r \sigma_k (\phi_k)_i (w_k)_j$  \\
    stand.~DeepONet & $\left(TB^T \right)_{ij}$                                           & $=$ & $\sum_{k=1}^N t_k(x_i) b_k(p_j)$            \\
    mod.~DeepONet   & $\left(\Phi_1 \Sigma_1 B^T \right)_{ij}$                            & $=$ & $\sum_{k=1}^N \sigma_k (\phi_k)_i b_k(p_j)$ \\
    \bottomrule
    \caption{Comparison of true solution (train/test), modified DeepONet and standard DeepONet in matrix formulation and for individual evaluation at coordinate $x_i$ and initial condition $p_j$. The rank of the data matrix is denoted as $r$. For better comparison to the modified DeepONet, the SVD of the true solution is displayed. Here, $W := (\Sigma^+ \Phi^T A_{te})^T$ contains the optimal coefficients for the test data; see~\cref{app:errordecompose_test_derive} for details.}
\end{longtable}

In this section, we modify the DeepONet by replacing the trunk network by the SVD basis. This modified DeepONet is used for the remainder of the work to investigate the branch network's error.
This facilitates the analysis significantly, since it reduces the DeepONet parameters to the branch network parameters.

As discussed in~\cref{sec:trunktheory}, for any full rank matrix $C$, the trunk matrix $T=\Phi_1 C$, with $\Phi_1$ chosen as the left singular vectors, minimizes the trunk error, which we call the SVD basis. We investigate this case here, using $C=\Sigma_1$ as a scaling, which corresponds to $t_k = \sigma_k \phi_k$ for the $k$-th output neuron. We motivate this choice in~\cref{app:trunksigmamotiv}.
Furthermore,~\cref{app:trunk_phi_vs_normal} contains results
for $T=\Phi_1$ (which coincides, up to a whitening of the input data,
with the POD-DeepONet~\cite{Lu_POD}) and a discussion of the impact of
the specific choice of $T$.

\Cref{tab:don_notation} compares the true solutions for the training and test data with the modified and standard DeepONets, both in matrix form (center column) and for individual coordinates and input functions (right column).
Recall that both $\Phi_1$ (the left singular vectors used as the trunk basis) and $V_1$ are obtained from the SVD of the training data matrix $A_{tr}$. The table
shows that the optimal branch matrix for the modified DeepONet evaluated on the training data is $V_1$, i.e., the matrix containing the right singular vectors.
As this will be relevant in~\cref{sec:spectralbias}, it is important to note that this is equivalent to stating that, for the initial condition $p_j$, the $k$-th branch output neuron is trained to predict $V_{jk}$.
The neural network architecture of the modified DeepONet is visualized in~\cref{fig:graphical_abstract}~(bottom left).

For the sake of simplicity in notation, we use the term DeepONet to refer to the modified DeepONet (with the SVD basis scaled with the singular values) for the remainder of this work.
When we use the term \textit{standard DeepONet}, we mean a DeepONet with a learned trunk basis.
Additionally, we call the left-singular vectors $\phi_i$ the \textit{modes} from now on.

\subsection{Mode Decomposition of the Branch Error}
\label{sec:modelossdecomp-deriv}

We decompose the (modified) DeepONet's training error by applying~\cref{eq:myerrordecomp} and, using the SVD of the training data matrix $A_{tr}$ and the resulting orthogonality, split it into trunk and branch components.
The branch error, which measures the total coefficient approximation error, then further decomposes into individual mode losses, each capturing how well a single mode's coefficient is approximated:
\begin{align}
    A_{tr}           & = \Phi_1 \Sigma_1 V_1^T + \Phi_2 \Sigma_2 V_2^T \nonumber                                                                                                                                                                    \\
    \varepsilon_{tr} & = ||\tilde{A}_{tr}-A_{tr}||_F^2 = || \Phi_1 \Sigma_1 B_{tr}^T - \Phi_1 \Sigma_1 V_1^T - \Phi_2 \Sigma_2 V_2^T ||_F^2 = || \Phi_1 \Sigma_1 B_{tr}^T - \Phi_1 \Sigma_1 V_1^T||_F^2 + || \Phi_2 \Sigma_2 V_2^T ||_F^2 \nonumber \\
                     & = || \Sigma_1 B_{tr}^T - \Sigma_1 V_1^T ||_F^2 + ||\Sigma_2||_F^2 =  \sum_{i=1}^N \sigma_i^2 \underbrace{|| b_{i, tr}-v_i ||_2^2}_{=: L_{i, tr}} + ||\Sigma_2||_F^2 \label{eq:modelossdecompose_train}
\end{align}
By design, the trunk space is spanned by the first $N$ left-singular vectors of $A_{tr}$, so $\varepsilon_T = ||\Sigma_2||_F^2$. With this choice the DeepONet achieves the optimal trunk error for the training data.
Since $||\Sigma_2||_F^2$ equals the trunk error, $||\Sigma_1 B_{tr}^T - \Sigma_1 V_1^T ||_F^2$ is the branch error.

\Cref{eq:modelossdecompose_train} thus shows the modification's main advantage for our analysis: the branch error decomposes into errors $\sigma_i^2 L_i$ for different modes.
The \textit{unweighted mode loss} $L_i$ is the difference between $v_i$ and $b_i$, that is, the true coefficients of mode $i$ and the coefficients $b_i$ predicted by the DeepONet, for all input functions.
Since the unweighted mode losses $L_i$ are scaled by the squared singular values $\sigma_i^2$ to yield the total error $\varepsilon$ in~\cref{eq:modelossdecompose_train}, we define the \textit{weighted mode loss} as $\sigma_i^2 L_i$.

A similar error decomposition can be derived for the test data. With $w_i$ denoting the optimal coefficients to approximate the test data using the training SVD basis, as detailed in~\cref{app:errordecompose_test_derive}, the test branch error decomposes as
\begin{align*}
    \varepsilon_{B,te} & = \sum_{i=1}^N \sigma_i^2 ||b_{i,te}-w_i||_2^2.
\end{align*}
To facilitate comparison between training and test mode losses, we define the test mode loss
\begin{align}
    L_{i,te} &:= \frac{m_{tr}}{m_{te}} ||b_{i,te} - w_i||_2^2 \label{eq:Li_test}
\end{align}
with a normalization factor such that both losses are scaled consistently; the precise definition is given in~\cref{app:errordecompose_test_derive}.

\begin{remark}
    To better judge whether a mode loss is low or high, we introduce the \textit{base loss}. We seek to define the base loss, such that observing a mode loss $L_i$ that exceeds its base loss indicates that the branch network is neglecting mode $i$.
    To this end, the base loss of mode $i$ is defined as the loss for $b_i=0$, i.e., when the DeepONet predicts a zero coefficient for mode $i$.
    On the training data, the unweighted base loss is $||v_i||_2^2 = 1$ because $V_1$ is semi-orthogonal, thus the weighted base loss is $\sigma_i^2$.
    For the test data, the unweighted base loss is $\frac{m_{tr}}{m_{te}} ||w_i||_2^2 \approx 1$.
    Thus, the weighted base loss is $\sigma_i^2 \frac{m_{tr}}{m_{te}} ||w_i||_2^2$.
    Observing a mode loss $L_i$ that exceeds its base loss means that the model performs worse than the trivial prediction $b_i=0$.
\end{remark}

\subsection{Distribution of Mode Losses}
\label{sec:modelossdistr}

In this section, we investigate how training and test branch errors distribute across modes.
We investigate three training approaches: plain GD (\cref{sec:gd}), re-weighted GD (\cref{sec:reweighting}), and Adam (\cref{sec:adam_training}), to understand their effects on mode-specific approximation quality.

Throughout this analysis, we use the following terminology: Since the $i$-th mode is associated with singular value $\sigma_i$ (ordered descendingly), we refer to modes with low indices as \textit{large modes}, modes with high indices as \textit{small modes}, and those in between as \textit{intermediate modes}.
We define large modes to have $\sigma_i^2/\sigma_1^2 > 0.1$ and small modes to have $\sigma_i^2/\sigma_1^2 < 10^{-3}$; this choice is somewhat arbitrary but the motivation for it will become clearer throughout the discussion of the results. For the standard example problem, this leads to large modes being $i \lesssim 10$ and small modes being $i \gtrsim 30$.
Note that we find qualitatively similar results for (a) the KdV equation with $T=\Phi_1$ in~\cref{app:trunk_phi_vs_normal}, and (b) the 2D heat equation with different input functions and $T=\Phi_1 \Sigma_1$ in~\cref{app:2d_modes}, except that the slower singular-value decay in 2D shifts the dominant contribution to the large modes.
\subsubsection{Gradient Descent Training}
\label{sec:gd}

\begin{figure}
    \centering
    \includegraphics{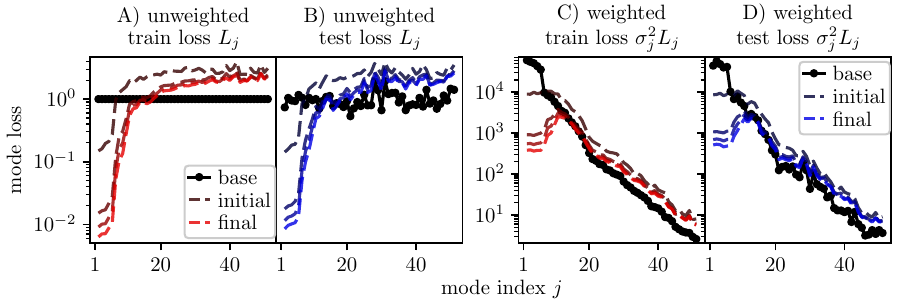}
    \caption{
        \textbf{Weighted mode losses for DeepONet trained with GD.}
        Unweighted (A,B) and weighted (C,D) training (A,C) and test (B,D) mode losses at different training stages, colored from gray (initial) to red/blue (final) shown as dashed lines. The respective base losses are shown as black dot-solid lines. The terms large, intermediate, and small modes refer to the singular-value regimes defined in~\cref{sec:modelossdistr}.
        For visual clarity, this figure contains no shaded bands.
    }
    \label{fig:modeerrors_GD}
\end{figure}

We first examine DeepONets trained using plain GD.
\Cref{fig:modeerrors_GD} shows both the unweighted mode losses $L_i$ and the weighted mode losses $\sigma_i^2 L_i$ for both training and test data over the course of training.
The respective base losses, defined in~\cref{sec:modelossdecomp-deriv}, are shown for reference.

An interesting pattern emerges: only the largest $\approx 10$ modes show significant loss reduction below their base losses during training.
This creates a characteristic error distribution in the weighted mode loss, where intermediate modes (indices 12-17 in our examples) contribute most to the total error; they are neither well-approximated like large modes nor negligible like small modes due to tiny singular values.
Note that this interior maximum is produced by training, not
by the $\sigma_i^2$ weighting alone: the peak emerges only because $L_i$ rises from $\approx 10^{-2}$ to $\approx 1$ over the same index range.
Furthermore, the mode losses corresponding to small singular values exceed the base losses, indicating neglect through the GD optimizer.
These learning dynamics are driven by gradient domination from large singular values. Since the gradient of the total loss with respect to the parameters is:
\begin{align*}
    \nabla_\theta \mathcal{L} = \frac{1}{nm} \sum_{i=1}^N \sigma_i^2 \nabla_\theta L_i,
\end{align*}
the contributions from small modes become negligible due to the $\sigma_i^2$ weighting, preventing the GD optimizer from improving their approximation even when their unweighted losses $L_i$ are large.

\begin{remark}
    A related phenomenon termed gradient starvation~\cite{gradientstarvation1, gradientstarvation2} has been identified in classification with cross-entropy loss, where correctly classified samples cease contributing to the gradient, causing dominant features to prevent the learning of other informative features. While our setting differs (we use MSE loss where all samples continue contributing), the common thread is that gradient-based optimization can systematically neglect certain learnable components when others dominate the gradient signal.
\end{remark}

\subsubsection{Modified Loss Re-Weighting}
\label{sec:reweighting}

To
further analyze the origin of
the gradient imbalance, we introduce a modified loss function with adjustable mode weighting:
\begin{align*}
    \mathcal{L}_{e,\text{tr}} = \frac{1}{nm_{\text{tr}}} \sum_{i=1}^N \sigma_i^{2+2e} L_{i,\text{tr}},
\end{align*}
where the exponent $e$ controls the re-weighting: $e < 0$ amplifies small modes, $e > 0$ emphasizes large modes, and $e = 0$ recovers the standard loss. To prevent gradient explosion or vanishing, we scale the learning rate by $\sigma_1^{-2e}$, keeping the largest mode's contribution invariant.
We use $\mathcal{L}_{e,\text{tr}}$ only to compute gradients (and thus parameter updates), not as a performance metric.
Note that the re-weighted loss for $e=-1$ is equivalent to using $\varepsilon_C$, defined in~\cref{eq:epsc}, instead of $\varepsilon_B$ in the loss function.

\begin{figure}[!htb]
    \centering
    \includegraphics{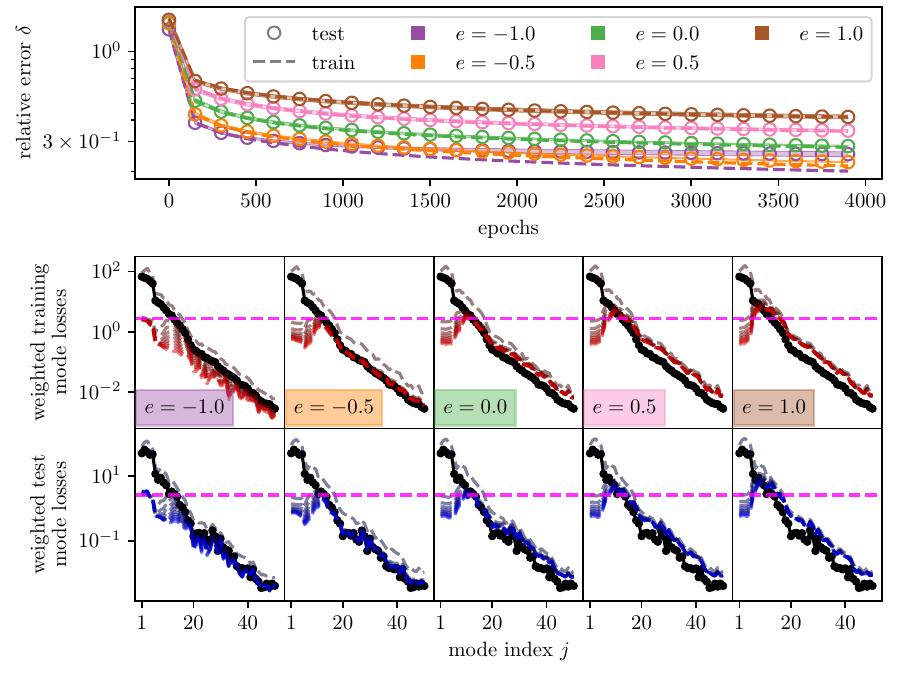}
    \caption{
        \textbf{Model performance across different exponents $e$ and training epochs for DeepONets trained using GD.}
        \textbf{Top panel:} Relative error $\delta = ||A- \Tilde{A}||_F/||A||_F$ for both training (dashed lines) and test (circles) data across different exponents ($e = -1.0, -0.5, 0.0, 0.5, 1.0$) over $\num{4000}$ epochs.
        \textbf{Center and bottom row:}
        Weighted training (center row) and test (bottom row) mode losses at different training stages, colored from gray (initial) to red/blue (final).
        Each column corresponds to a different exponent $e$. The third column shows the DeepONet trained using the standard loss ($e=0$).
        The plots in the center and bottom row contain the respective base losses in black, and a pink dashed horizontal line marking the maximum mode loss in the last training epoch of $e=0$, facilitating comparison between different exponents $e$.
        For visual clarity, the training-loss curve in the top panel and the plots in the center and bottom rows have no shaded bands.
    }
    \label{fig:modeerrors_weighting_GD}
\end{figure}

\Cref{fig:modeerrors_weighting_GD} shows the effect of this re-weighting. For $e = -1$, which weights all modes equally regardless of their singular values, we achieve significant loss reduction across all training modes. However, this comes with a critical trade-off: small modes improve dramatically in training but severely overfit, with test losses near base levels despite low training losses.
In our experiments, the optimal balance emerges at $e = -0.5$, which achieves the lowest test error for two reasons.
First, it maintains reasonable approximation quality for large modes (better than $e = -1$).
Second, it meaningfully improves intermediate and small modes (better than $e \geq 0$).
For $e > 0$, the optimizer neglects small modes, with their losses exceeding base levels, i.e., higher loss than zero coefficients. Only the first 10, 8, and 5 modes improve for $e = 0.0, 0.5,$ and $1.0$ respectively, showing how increased emphasis on large modes comes at the expense of all others.

This analysis reveals a fundamental insight: the standard loss function ($e = 0$) is poorly calibrated for the multi-scale nature of the mode decomposition. The steep singular value decay creates an optimization landscape where gradients are dominated by a small subset of modes, leaving most of the representation capacity unused.
We introduce this re-weighting primarily for analysis; although it may be used in practice, we do not propose it here as a practical algorithm.

\subsubsection{Adam Optimizer}
\label{sec:adam_training}

\begin{figure}
    \centering
    \includegraphics{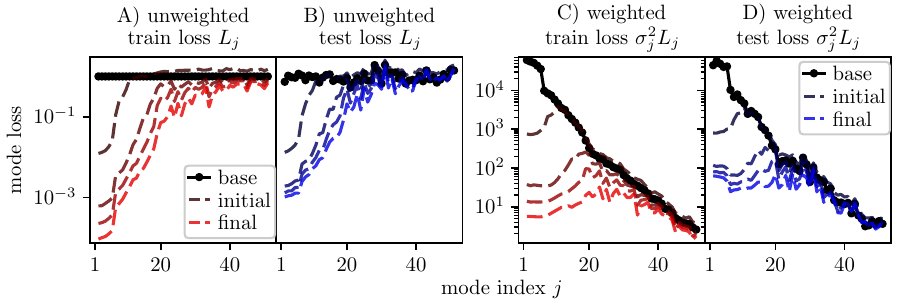}
    \caption{
        \textbf{Weighted mode losses for DeepONet trained with Adam.}
        Unweighted (A,B) and weighted (C,D) training (A,C) and test (B,D) mode losses at different training stages, colored from gray (initial) to red/blue (final) shown as dashed lines. The respective base losses are shown as black dot-solid lines.
        For visual clarity, this figure contains no shaded bands.
    }
    \label{fig:modeerrors_Adam}
\end{figure}

The Adam optimizer provides an alternative to the gradient imbalance problem through its adaptive learning rate mechanism.
\Cref{fig:modeerrors_Adam} shows that Adam achieves significantly lower total losses than GD with standard weighting, and its mode loss distribution is more similar to GD with $e \approx -0.5$ or $-1.0$.

Unlike GD with $e = 0$, Adam reduces all training mode losses below their base values, even for small modes, though at different rates. The optimizer achieves significant loss reduction on approximately 30 modes for both training and test data, whereas GD effectively improves only the first ten. This suggests that Adam's per-parameter learning rate adaptation compensates for the singular value imbalance, effectively implementing an implicit re-weighting scheme.

The mode loss distribution under Adam can be characterized as follows:
Large modes ($i \lesssim 10$) are well-approximated with low weighted losses. Note that Adam approximates the large modes far better than GD; see~\cref{fig:modeerrors_GD}.
Small modes ($i \gtrsim 30$) contribute negligibly due to tiny singular values.
The intermediate modes' losses are significantly improved compared to base losses but still contribute significantly to the total error due to non-negligible singular values.
Furthermore, note that large modes have a significantly higher contribution to the test loss compared to the training loss.
When explicitly re-weighting Adam ($e \neq 0$), we observe that $e = -1.0$ yields approximately homogeneous unweighted mode losses $L_{i,\text{tr}} \approx \text{constant}$ for the training data; this can be inferred from the weighted mode losses in~\cref{app:adam_reweight}, since the weighting factor $\sigma_i^2$ is known.
However, similar to GD, aggressive emphasis on small modes ($e < 0$) leads to overfitting.
For Adam, we observe optimal test performance at $e = 0$.
This indicates that Adam's implicit re-weighting already strikes a good balance for generalization.

\begin{remark}
    The superior performance of adaptive gradient methods extends beyond Adam. For instance, AdaGrad shows similar improvements over standard GD; see~\cref{app:adagrad}. This consistent pattern suggests that addressing the gradient imbalance through adaptive learning rates is crucial for operator learning tasks where the loss function admits a multi-scale decomposition.
\end{remark}

\begin{remark} \label{remark:choose_N_according_to_Li}
A mode with $L_i \approx 1$ contributes $\sigma_i^2$ to the branch error, which is exactly what it would contribute to the trunk error if it were truncated instead.
Increasing $N$ to include modes that the branch network does not actually learn thus leaves the total error essentially unchanged, and the mode loss distribution indicates whether $N$ or the branch network's training is the limiting factor.
\end{remark}
\begin{remark} \label{remark:mode_lens_vs_aggregate}
The mode loss decomposition reveals structure that the aggregate error curve conceals. The aggregate relative test error merely stagnates (\cref{fig:modeerrors_Ada}), while the mode losses identify its composition: under Adam, the gap between training and test error is produced almost entirely by overfitting the intermediate and small modes, whose training losses fall below the base loss while their test losses remain at it (\cref{fig:modeerrors_Adam}). Under GD, some modes are approximated worse than the trivial prediction $b_i=0$ (\cref{fig:modeerrors_GD}). Since a DeepONet with $T=\Phi_1$ coincides, up to a whitening of the input data, with the POD-DeepONet~\cite{Lu_POD}, the same analysis applies to that architecture; \cref{app:trunk_phi_vs_normal} shows that the mode loss distribution is qualitatively unchanged.
\end{remark}

\subsection{Mechanisms Influencing the Mode Loss Distribution}
\label{sec:why}

Having established how different optimizers affect the mode loss distribution, we now investigate the underlying mechanisms driving these patterns.
Beyond the weighting by $\sigma_j^2$, two factors can make the mode losses depend on the mode index $j$: the functions the branch network must learn may differ in difficulty across modes, and all mode coefficients are produced by one shared network, so the modes are not learned
independently. We therefore examine three complementary perspectives: spectral bias (\cref{sec:spectralbias}), which asks whether the mode losses follow the frequency content
of the corresponding branch targets; architectural coupling (\cref{sec:stacked}), which contrasts the shared branch network with the stacked DeepONet, where each coefficient has
its own network; and update-based coupling (\cref{sec:coupling}), which quantifies how
strongly a gradient step reducing one mode's loss perturbs the others; the modes are orthogonal in output space, but entangled in parameter space.
\subsubsection{Spectral Bias}
\label{sec:spectralbias}

\textit{Spectral bias}~\cite{rahaman-sb, xu-sb} is the tendency of neural networks to learn low-frequency components faster and more accurately than high-frequency ones. We investigate whether this affects the branch network's mode loss distribution.

Since~\cref{sec:trunk} shows trunk error is negligible for large $N$, we focus on the branch network. The modification in~\cref{sec:pod_don} provides well-defined target functions for analysis: the $j$-th branch neuron predicts $V_{kj}$ for input $p_k$, defining the \emph{right singular function} $\rho_j : p_k \mapsto V_{kj}$.
If spectral bias affects the branch, mode losses $L_j$ should correlate with $\rho_j$ frequency: high-frequency $\rho_j$ should yield higher losses.

\begin{figure}[!ht]
    \centering
    \includegraphics{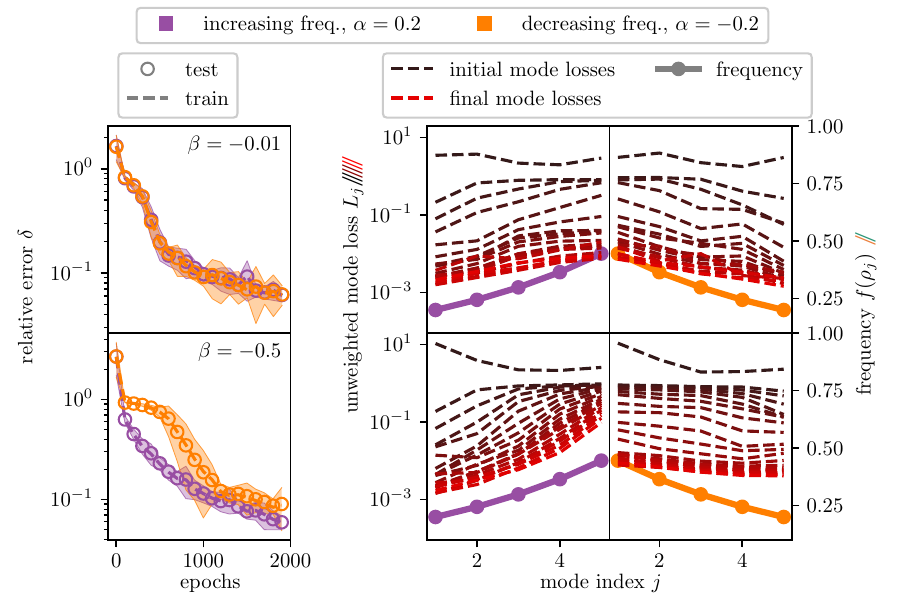}
    \caption{
        \textbf{Spectral bias in modified DeepONet for synthetic data.}
        \textbf{Left:} Relative error over epochs (training error as dashed line and test error as circles; nearly identical).
        \textbf{Center/right:} Unweighted mode losses (left scale) at different training stages (colored from black (initial) to red (final)) with frequencies of right singular functions (right scale).
        The center column shows the case of increasing frequencies ($\alpha=0.2$), and the right column shows the case of decreasing frequencies ($\alpha=-0.2$); the corresponding relative errors are shown in purple and orange respectively in the left column.
        \textbf{Top/bottom:} The top row shows synthetic data with singular values $\sigma_j = \exp(-0.01 j)$, while the bottom row has $\sigma_j = \exp(-0.5 j)$. Each curve is averaged over 20 randomly initialized DeepONets.
        For visual clarity, the center and right plots contain no shaded bands.
    }
    \label{fig:sb_synth}
\end{figure}

We define the function frequency of $g : \mathbb{R}^d \to \mathbb{R}$ using the Fourier transform $\mathcal{F}(g)(\xi)  = \int_{\mathbb{R}^d} g(x) \exp(-2 \pi i \xi^T x) dx$.
The mean frequency~\cite{mallat} is $f(g)  = \frac{\int_{\mathbb{R}^d} |\mathcal{F}(g)(\xi)|^2 ||\xi||_2 d\xi}{\int_{\mathbb{R}^d} |\mathcal{F}(g)(\xi)|^2 d\xi}$.
We first investigate spectral bias in the branch network on a synthetic dataset with curated spectral properties.
As described in more detail in~\cref{app:syntheticdata}, the synthetic data is created such that the mean frequency of $\rho_j$ is
\begin{align*}
    f(\rho_j) & \approx
    \begin{cases}
        F_0 \exp(\alpha (j-1))   & \text{ if } \alpha > 0, \\
        F_0 \exp(\alpha (j-1-N)) & \text{ else}.
    \end{cases}
\end{align*}
Thus, $F_0$ denotes the minimum frequency and $\alpha$ denotes the rate of increase (or decrease) of the frequencies. Here, we use $F_0 = 0.2$.
Furthermore, we construct the synthetic data so that its singular values satisfy $\sigma_j = \exp(\beta j)$.
Since $\beta = 0$ makes the ordering of right singular vectors non-unique, we require $\beta \neq 0$.
Firstly, we consider $|\beta|$ small to isolate spectral bias effects from singular value weighting.
For instance, $\beta = -0.01$ in~\cref{fig:sb_synth}~(top) yields $\sigma_1^2 / \sigma_5^2 \approx 1.08$, so all modes are approximately equally relevant while remaining correctly ordered.
We then compare two cases: (i) $\alpha = 0.2$ and (ii) $\alpha = -0.2$.
The mode losses in both cases are very similar, and in both cases, the low-frequency modes have smaller mode losses.
In~\cref{fig:sb_synth}~(bottom), we consider the same frequencies $\alpha=\pm 0.2$, but $\beta=-0.5$, i.e., a strong prioritization of the first few modes ($\sigma_1^2 \approx 55 \sigma_5^2$).
Here, we see that despite the initially slower learning of the higher-frequency features, they are eventually learned due to their higher relevance. This illustrates the competing effects of loss weighting and spectral bias.

\begin{figure}[!ht]
    \centering
    \includegraphics{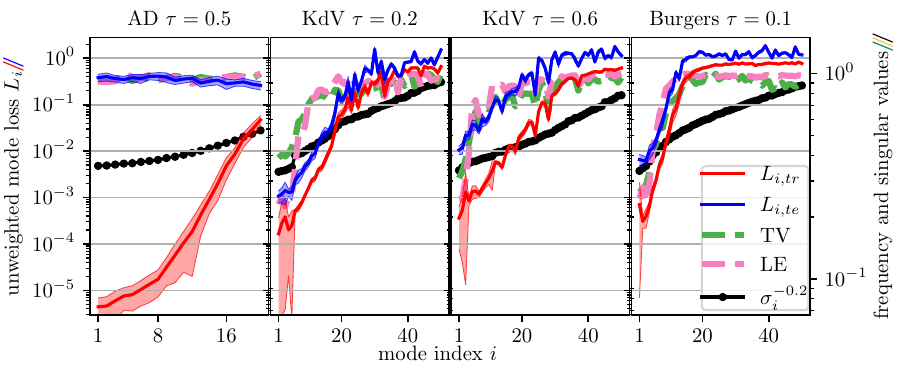}
    \caption{
        \textbf{Spectral bias in modified DeepONet for different PDEs.}
        Unweighted training (red) and test (blue) mode losses (left scale),
        the frequencies (right scale) of the right singular functions $\rho_i$ estimated via the TV norm with $k=3$ (green) and LE $k=50$ (pink)
        and the scaled singular values $\sigma_i^{-0.2}$ (black, right scale) are shown.
        \textbf{Example problems (by column):} advection-diffusion (AD) equation with $\tau=0.5$, KdV equation with $\tau=0.2$, KdV equation with $\tau=0.6$ and Burgers' equation with $\tau=0.1$.
    }
    \label{fig:spectralbias_branch}
\end{figure}

We now return to the PDE datasets (advection-diffusion, KdV, and Burgers) and investigate the impact of spectral bias on the mode loss distribution.
The (inverse) scaled singular values $1/\sigma_i^{0.2}$, the unweighted mode losses, and the frequencies $f(\rho_i)$ of the right singular functions $\rho_i$ are shown in~\cref{fig:spectralbias_branch} for the following considered example problems: advection-diffusion with $\tau=0.5$, KdV with $\tau=0.2$, KdV with $\tau=0.6$, and Burgers with $\tau=0.1$.
To show the trend of the frequencies of $\rho_i$, we choose two methods to estimate the frequencies:
the total variation (TV) norm and the Laplacian energy (LE).
The TV norm estimates frequency by measuring how rapidly the function values change between neighboring samples, while the LE uses concepts from spectral graph theory to quantify oscillations on a graph constructed from the data. Both methods are $k$-nearest neighbor methods; we use $k=3$ and $k=50$ for the TV norm and LE, respectively. A detailed description of both methods can be found in~\cref{app:frequencyestimation}.
Both estimates are heuristic: we are not aware of a reliable method for quantifying the frequency of a multivariate function sampled at scattered points; see~\cref{app:frequencyestimation}.
On the synthetic data, both reproduce the prescribed frequency ordering (\cref{sec:fourier_closing}), but for the PDE data, we read them only as qualitative trend indicators within a single problem.
Note that each plot in~\cref{fig:spectralbias_branch} contains two $y$-axis scales, left and right, both logarithmic.

For the KdV and Burgers equations (second--fourth columns in~\cref{fig:spectralbias_branch}), we observe a clear correlation between $f(\rho_i)$ and the mode losses $L_i$ for both test and training data.
Since both test and training mode losses are low for low frequencies and high for high frequencies, the branch network learns a generalizing approximation for low frequencies while underfitting high frequencies.

For the advection-diffusion equation (first column), the estimated frequencies $f(\rho_i)$ vary only weakly with $i$. Unlike the KdV and Burgers equations, no general trend is observable.
However, we observe an approximately constant test mode loss $L_{i,te}$, indicating significant overfitting.
Due to the different input dimensions $M$ across the example problems, a reliable method to compare the frequencies of the $\rho_i$ between problems is not available.
Consequently, an explanation based on spectral bias for why the DeepONet overfits the advection-diffusion equation compared to other equations cannot be given.
An alternative hypothesis, unrelated to spectral bias, is that the advection-diffusion problem's low rank (exactly 20, matching the inner dimension $N$) removes the approximation pressure present in higher-rank problems like KdV, potentially allowing the network to memorize training data without learning generalizable structure, thus resulting in uniform overfitting across all modes. However, confirming this hypothesis requires further investigation in future work.

\subsubsection{Architectural Coupling}
\label{sec:stacked}

\begin{figure}[!ht]
    \centering
    \begin{minipage}[c]{0.48\textwidth}
        \begin{subfigure}[b]{\textwidth}
            \centering
            \begin{tikzpicture}
            \node[anchor=south west, inner sep=0] (img)
              {\includegraphics[width=1.0\textwidth]{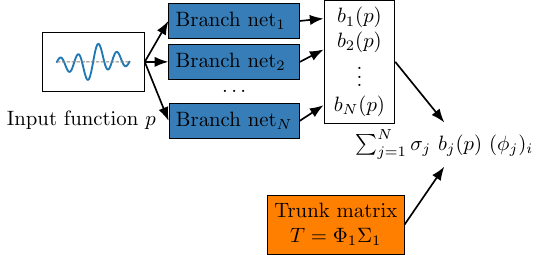}};
            \node[anchor=south west, fill=white, inner sep=2pt]
              at (img.south west) {(a)};
           \end{tikzpicture}
           \phantomcaption\label{fig:stacked_svd_scheme}
        \end{subfigure}
        \begin{subfigure}[b]{\textwidth}
            \centering
            \begin{tikzpicture}
            \node[anchor=south west, inner sep=0] (img)
            {\includegraphics{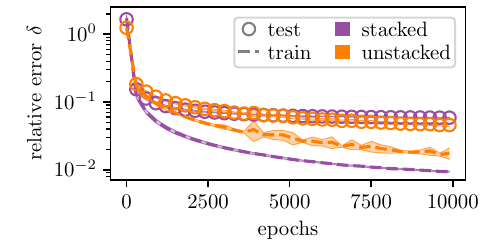}};
            \node[anchor=south west, fill=white, inner sep=2pt]
              at ([xshift=5pt, yshift=0pt]img.south west) {(b)};
           \end{tikzpicture}
           \phantomcaption\label{fig:stacked_losscurves}
        \end{subfigure}
    \end{minipage}
    \hfill
    \begin{minipage}[c]{0.48\textwidth}
        \vspace{0.8cm}
        \begin{subfigure}[b]{0.49\textwidth}
            \centering
            \begin{tikzpicture}
            \node[anchor=south west, inner sep=0] (img)
            {\includegraphics{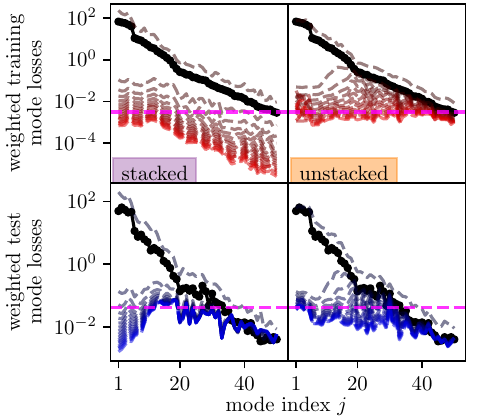}};
            \node[anchor=south west, fill=white, inner sep=2pt]
              at ([xshift=0pt, yshift=0pt]img.south west) {(c)};
           \end{tikzpicture}
           \phantomcaption\label{fig:stacked_modelosses}
        \end{subfigure}
    \end{minipage}
    \caption{
        \textbf{Stacked and unstacked DeepONets.}
        \textbf{(a) Architecture of the stacked modified DeepONet.}
        \Cref{fig:graphical_abstract} (bottom left) shows the unstacked DeepONets for reference.
        Adapted and reproduced from~\cite{deeponet}.
        \textbf{(b) Relative errors for stacked and unstacked DeepONets trained with Adam.}
        \textbf{(c) Weighted mode losses for stacked and unstacked DeepONets trained with Adam.}
        Weighted training (top row) and test (bottom row) mode losses at different training steps, colored from gray (initial) to red/blue (final).
        Columns correspond to different architectures and hidden layer widths (left: stacked with $w_{sta}=42$, right: unstacked with $w_{unst}=495$), as indicated by the labels.
        The top and bottom row plots also contain the respective base losses in black, and a pink dashed horizontal line marking the maximum mode loss of the fully-trained unstacked DeepONet.
        For visual clarity, figure (c) contains no shaded bands.
    }
    \label{fig:stacked_vs_unstacked_all}
\end{figure}
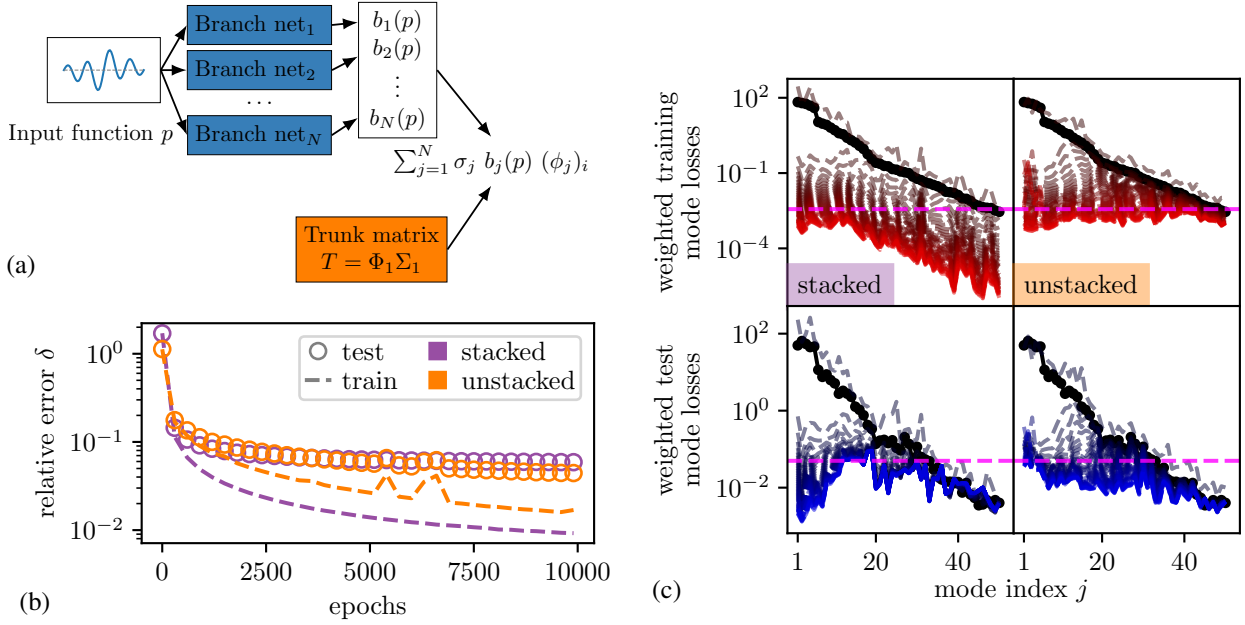

Both the standard and the modified DeepONet,~\cref{sec:pod_don}, use a single branch network with shared hidden layers outputting all mode coefficients simultaneously.
However, together with the normal (or unstacked) DeepONet, Lu et al.\ also proposed the so-called stacked DeepONet~\cite{deeponet}.
Instead of one multi-layer perceptron with $N$ output neurons, the branch network in the stacked DeepONet consists of $N$ multi-layer perceptrons with one output neuron each, see~\cref{fig:stacked_svd_scheme}. We thus say that the coefficients are \textit{architecturally coupled} in the unstacked DeepONet.

To investigate the influence of architectural coupling, we compare the performance of stacked and unstacked DeepONets.
Specifically, we use a stacked DeepONet whose branch networks each have depth $D=5$ and hidden layer width $w_{sta}=42$.
We compare this to an unstacked DeepONet with the same number of trainable parameters. To achieve this, we choose $w_{unst} = 495$ and $D=5$.

\Cref{fig:stacked_losscurves} and~\ref{fig:stacked_modelosses} show the total loss over training and the mode losses of the stacked and unstacked DeepONets, respectively.
The stacked DeepONet achieves a lower training error, while the unstacked DeepONet achieves a slightly lower test error, indicating overfitting in the stacked DeepONet.
The mode losses show that the stacked DeepONet does not overfit uniformly across modes.
For large modes, the stacked DeepONet achieves the smallest test and training errors.
However, for small modes, the stacked DeepONet's low training losses do not transfer to the test data.
Even for intermediate modes, the unstacked network generalizes significantly better than the stacked DeepONet.

The first key observation from~\cref{sec:modelossdistr} was that small modes seem to generalize poorly when fitted well on training data. The stacked DeepONet experiments corroborate this hypothesis.
Furthermore, we observe that unstacking, or neuron-sharing, helps overall generalizability but not for all modes.
In particular, it seems to negatively impact the training loss and generalizability of large modes while improving the generalizability of smaller modes.

\begin{remark}
    This counterintuitive result connects to insights from MTL~\cite{ruder2017overviewmultitasklearningdeep}.
    Recall that the shared hidden layers in the unstacked architecture force the network to learn representations that work across multiple modes. When different tasks (here, approximating different mode coefficients) share an underlying structure, this parameter sharing acts as an implicit regularizer, promoting sets of parameters that generalize well rather than overfitting to individual modes.
    This implies that even when practitioners require only specific modes (e.g., the coefficients for specific spatial frequencies), the unstacked architecture may still be preferable. The parameter sharing facilitates learning of generalizable features that improve approximation quality for target modes, even if other outputs are discarded.
    Note that this is a different connection to MTL than works such as~\cite{poseidon, synergy, zhang2024deeponetmultioperatorextrapolationmodel} draw; they train a so-called foundation model on different PDEs, while we consider multiple tasks within one PDE.
\end{remark}

\subsubsection{Update-Based Coupling}
\label{sec:coupling}

In comparing stacked and unstacked DeepONets, we investigated how rearranging the branch network so that modes have separated parameters (while keeping the number of parameters constant) affects generalization.
In this section, we consider unstacked DeepONets and investigate further how much of the loss stems from mode coupling.
Changing the value of the $i$-th branch output neuron does not change the contribution of any other mode to the DeepONet output because of the modes' orthogonality ($\phi_i^T \phi_j = 0$ for $i \neq j$).
However, when training with gradient descent to reduce the loss for mode $i$, this parameter update changes all network parameters, since they are shared between modes. Thus, the mode $i$-specific parameter update can significantly alter the values of all branch output neurons, not just the $i$-th one, due to the neuron-sharing discussed in~\cref{sec:stacked}.
We study the modes' coupling in parameter space to understand when learning one mode helps to learn the others (\textit{beneficial coupling}) and when the modes get in each other's way (\textit{detrimental coupling}).

We now give a formal definition of \textit{mode coupling in parameter space} for DeepONets trained with GD.
Unless stated otherwise, any loss function in this section is evaluated at the current parameters $\theta$, and gradients are computed with respect to $\theta$.
For clarity, we write $\nabla L_i$ instead of $\nabla_\theta L_i(\theta)$, and $\nabla \mathcal{L}$ instead of $\nabla_\theta \mathcal{L}(\theta)$.
Consider a GD parameter update
\begin{align*}
    \delta \theta & = -\alpha \nabla \mathcal{L}_{tr} = -\frac{\alpha}{n m_{tr}}\sum_{j=1}^N \sigma_j^2 \nabla L_{j, tr},
\end{align*}
where we also drop the iteration index $t$. Choosing a sufficiently small learning rate $\alpha$ allows us to approximate the change in the training and test loss function using a first-order Taylor expansion.
As we will see, the Taylor expansion allows us to partition the loss change into coupling and non-coupling parts.
The first-order Taylor expansion of the total loss (train or test) change $\Delta \mathcal{L}$ is
\begin{align*}
    \Delta \mathcal{L} & \approx (\nabla \mathcal{L})^T \delta \theta = \frac{1}{n m_{tr}} \sum_{i=1}^N \sigma_i^2 (\nabla L_i)^T \delta \theta = -\frac{\alpha}{n^2 m_{tr}^2} \sum_{i=1}^N \sum_{j=1}^N \sigma_i^2 \sigma_j^2 (\nabla L_i)^T \nabla L_{j, tr} \eqqcolon d + \Omega,
\end{align*}
where
\begin{align*}
    d & \eqqcolon - \frac{\alpha}{n^2 m_{tr}^2} \sum_{i=1}^N \sigma_i^4 (\nabla L_i)^T \nabla L_{i, tr}
    \qquad \text{and} \qquad
    \Omega \eqqcolon - \frac{\alpha}{n^2 m_{tr}^2} \sum_{i=1}^N \sum_{j\neq i} \sigma_i^2 \sigma_j^2 (\nabla L_i)^T \nabla L_{j, tr}.
\end{align*}
We call $d$ the \textit{diagonal term} and $\Omega$ the \textit{off-diagonal term}.
Note that $d_{tr} \leq 0$, by definition.
Then, we define the relative coupling strength as
\begin{align*}
    \gamma \coloneqq \frac{\Omega}{d + \Omega}.
\end{align*}
Assume successful training, i.e., $d+\Omega < 0$,
then detrimental coupling between the modes, i.e., $\Omega > 0$, is symbolized by $\gamma < 0$, while beneficial coupling $\Omega < 0$ is symbolized by $\gamma > 0$.
Furthermore, training and test performance of DeepONets trained with GD have so far been observed to be very similar; these DeepONets are underfitting.
We thus only explicitly focus on discussing the results for the training data.

\begin{figure}[!ht]
    \centering
    \begin{subfigure}[t]{0.4\textwidth}
        \includegraphics{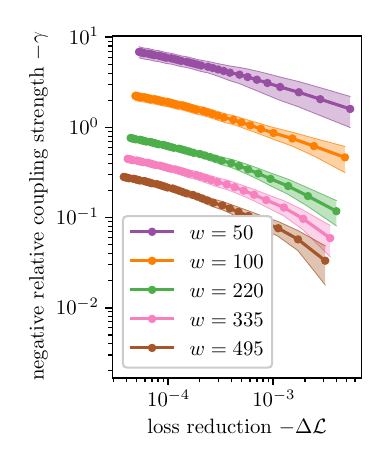}
        \caption{
            \textbf{Negative relative coupling strength $-\gamma$ plotted over loss reduction $-\Delta \mathcal{L}$ for DeepONets of different hidden-layer widths $w$.}
            The figure shows models with hidden layer widths 50 (purple), 100 (orange), 220 (green), 335 (pink) and 495 (brown).
        }
        \label{fig:couplingstrengths_and_width}
    \end{subfigure}
    \quad
    \begin{subfigure}[t]{0.56\textwidth}
        \includegraphics{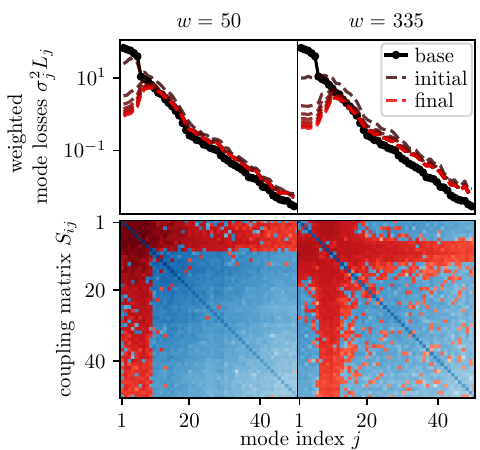}
        \caption{
            \textbf{Mode losses and coupling matrices.}
            \textbf{Top row:} Weighted mode losses at different training stages, colored from gray (initial) to red (final). Singular values shown in black.
            \textbf{Bottom row:} Entries of the $S_{tr}$ matrix after $\num{4000}$ epochs. Red indicates a positive entry (detrimental coupling) and blue indicates a negative entry (beneficial coupling). Both matrices use the same color scale.
            \textbf{Left / right:} Hidden layer width $w=50$ / $w=335$.
            For visual clarity, this figure contains no shaded bands.
        }
        \label{fig:couplmat}
    \end{subfigure}
    \caption{\textbf{Update-based coupling for DeepONets of different hidden-layer widths.}}
    \label{fig:coupling}
\end{figure}

We can now compute $\gamma$ over the course of the training for different DeepONets.
\Cref{fig:couplingstrengths_and_width} plots the (negative) relative coupling strength over the loss reduction for DeepONets of different hidden layer widths.
For all hidden layer widths, we find detrimental coupling, $\gamma < 0$.
Furthermore, we find strong decoupling with increased width; i.e., the wider DeepONets exhibit significantly smaller negative coupling strengths.
Note that these wider, decoupled DeepONets also achieve lower test and training losses.

Furthermore, we can inspect the coupling directly by defining the coupling matrix $S \in \mathbb{R}^{N \times N}$ with entries
\begin{align*}
    S_{ij} = -\frac{\alpha}{n^2 m_{tr}^2} \sigma_i^2 \sigma_j^2 (\nabla L_i)^T \nabla L_{j,tr}.
\end{align*}
Thus, $d+\Omega = \sum_{i,j=1}^N S_{ij}$.
\Cref{fig:couplmat} shows the mode losses and the coupling matrices for DeepONets with $w=50$ and $w=335$.

For $w=50$, most entries in the first ten rows and columns are positive.
We furthermore note that only the first ten modes' losses are significantly lower than their base losses; we thus refer to them (a bit euphemistically) as \textit{well-approximated}.
Thus, the coupling between a well-approximated mode and any other mode is likely to be detrimental.
However, the coupling between two poorly approximated modes ($i,j > 10$) is likely to be beneficial; $S_{ij} < 0$ (blue).
The matrix for $w=335$ shows a more complex pattern. Here, we find $S_{ij} < 0$ for $i<5,j>15$ and $j<5,i>15$, whereas $S_{ij} > 0$ for $5<i<15$ and $5<j<15$, and $S_{ij} < 0$ for $i>15,j>15$. Thus, the exact coupling mechanism seems to be dependent on hyperparameter choice.

\section{Conclusion and Discussion}
\label{sec:discussion}

\paragraph{Trunk and Branch Error in DeepONets}
We decompose DeepONet error into trunk and branch components, adapting Lanthaler et al.~\cite{Lanthaler_errorestimates_2022}. Across PDE examples, the branch network (learning basis coefficients) dominates approximation error for large inner dimensions, while the trunk learns a sufficient basis.
To the best of the authors' knowledge, this is the first empirical, systematic investigation of the distribution between trunk and branch errors.

\paragraph{Modified DeepONet with Fixed SVD Basis}
To further investigate the branch error, we construct a modified DeepONet with a fixed SVD basis and briefly compare it to the standard DeepONet.
We reiterate that this modified DeepONet is intended primarily as a diagnostic tool for analyzing the branch error in a controlled setting, rather than as a deployable architecture.
For simplicity, we assume the standard SVD algorithm for a full data matrix, that is, the data must be sampled on the same spatial grid for every parameter configuration.
This is computationally intensive and may not be feasible in higher-dimensional spatial domains, where the number of grid points grows exponentially with the spatial dimension (curse of dimensionality), but facilitates our analysis.
The standard DeepONet is less restrictive regarding its training data.
Furthermore, the modified DeepONet can only be evaluated at specific grid positions. If interpolation is used in the trunk, however, it can be evaluated at arbitrary positions, similar to the standard DeepONet.
We believe that our findings regarding the modified DeepONet are transferable to the standard DeepONet, since the standard DeepONet's trunk network learns to approximate the SVD basis.
The full-grid requirement could potentially be relaxed using low-rank matrix completion techniques (e.g., soft-thresholded singular value imputation~\cite{mazumder_softimpute} or nuclear-norm minimization~\cite{candes_recht}), which recover a low-rank factorization from partially observed matrices.
We note, however, that the associated recovery guarantees depend on the structure of the missing entries, and that the curse of dimensionality affects the generation and storage of the snapshots themselves, not only the SVD computation.
A thorough investigation of such a deployable variant is beyond the scope of the present work and left for future research.

\paragraph{Mode Decomposition of the Branch Error}
Mode-specific analysis (enabled by the DeepONet modification) reveals poor generalization for modes with small singular values.
Furthermore, distinct but characteristic mode loss distributions for DeepONets trained with GD and Adam can be observed. This highlights GD's failure to capture more than the largest few modes.
Three complementary effects shape this distribution:
For examples involving the KdV and Burgers equations, mode losses correlate with heuristic frequency
estimates of the right singular functions, consistent with spectral bias; the advection-diffusion example shows no such trend, and the impact on generalization remains unclear.
Architectural coupling (stacked vs. unstacked) produces different mode loss distributions and generalization, connecting operator learning to multi-task learning.
Update-based coupling shows network width reduces mode interactions.
These effects interact: right singular function frequency appears to affect
coefficient learnability, while parameter sharing in unstacked architectures further impacts approximation quality and training dynamics.

\paragraph{Practical Implications}
Network width is crucial for reducing mode coupling. MTL methods~\cite{gradconf1, gradconf2} for gradient conflicts may address mode coupling in DeepONets.
Adaptive optimizers (e.g., Adam) help learn beyond dominant modes, and unstacked architectures improve generalization through parameter sharing.
Training data SVD can inform hyperparameter choices like inner dimension; see remarks~\ref{remark:choose_N_based_on_sigma} and~\ref{remark:choose_N_according_to_Li}.
The modified DeepONet and mode-decomposition analysis provide tools for understanding and improving operator learning.

\paragraph{Limitations and Future Work}
While our analysis focuses on time evolution operators, the error decomposition framework is not inherently restricted to this setting and may extend to other operator types, such as steady-state solution operators, integral operators, and parameter-to-solution maps.
Open questions remain, in part due to our focus on relatively smooth PDE solutions.
It is not clear how our analysis would change when the trunk error becomes more pronounced, for instance, for hyperbolic equations with discontinuous initial conditions or shock formations.
Moreover, this work mainly considers low-dimensional spatial problems and mainly uses the same coordinates for testing and training. While the results in~\cref{app:offgrid} and~\cref{app:2d} support our claims, a more thorough investigation of more general settings is warranted.
Additionally, extending this analysis to more advanced DeepONet variants and architectures, such as those with Fourier features or Kolmogorov-Arnold networks, is an important direction for future work.
Lastly, to clarify the impact of spectral bias, future work should include more example problems with varied spectral properties and improved frequency estimation methods.

\section{CRediT authorship contribution statement}

\textbf{Alexander Heinlein}: Conceptualization, Methodology, Investigation, Writing - review and editing, Supervision \\
\textbf{Johannes Taraz}: Conceptualization, Methodology, Software, Validation, Investigation, Writing - original draft, Writing - review and editing, Visualization

\section{Declaration of competing interest}
The authors declare that they have no known competing financial interests or personal relationships that could have appeared to
influence the work reported in this paper.

\section{Code availability statement}
The implementation is available at \url{https://github.com/jotaraz/ModeDecomposition-DeepONets}.

\section{Acknowledgments}
We thank Henk Schuttelaars for conceptual discussions during the early stages of this work and for feedback on related drafts.

\section{Declaration of generative AI in the manuscript preparation process}

During the preparation of this work, the authors used Claude~\cite{claude} and Codex (OpenAI)~\cite{openai_codex} for proofreading and language editing. After using these tools, the authors reviewed and edited the content as needed and take full responsibility for the content of the published article.

\bibliographystyle{elsarticle-num}
\bibliography{references_formatted}

\appendix

\section{Appendix: Hyperparameters}
\label{app:hyperparams}

The hyperparameters used in this work are similar to those employed in related works~\cite{deeponet, Lu_POD, williams2024physicsinformeddeeponetslearnunderstanding}, balancing accuracy and cost. We do not perform extensive hyperparameter tuning: the aim is not to obtain particularly well-performing DeepONets, but to analyze the error structure of DeepONets with standard hyperparameter choices. Accordingly, no validation set is used, and the test data are never used to select hyperparameters.
Main findings (branch error dominance, mode loss distribution) remain qualitatively consistent across tested ranges: learning rates $[10^{-4}, 8 \times 10^{-3}]$, depths 3-10, widths 50-500, and different initializations.

The initial learning rate $\alpha_1 = 10^{-4}$, used throughout~\cref{sec:coupling}, is chosen such that the first-order Taylor expansion of the loss change approximates the true loss change $\Delta \mathcal{L}$.
        In addition to GD (\cref{eq:gd}), we use the Adam optimizer in this work.
 The update $\theta_{t+1} = \theta_t + \delta \theta_t$ uses:
\begin{align*}
    m_t &= \beta_1 m_{t-1} + (1-\beta_1) \nabla \mathcal{L}_{tr}(\theta_t),   &  v_t &= \beta_2 v_{t-1} + (1-\beta_2) (\nabla \mathcal{L}_{tr}(\theta_t))^2, \\
    \hat{m}_t       & = \frac{m_t}{1-\beta_1^t}, & \hat{v}_t       & = \frac{v_t}{1-\beta_2^t},  \\
    \delta \theta_t & = -\alpha_t \frac{\hat{m}_t}{\sqrt{\hat{v}_t + \bar{\epsilon}} + \epsilon}. & &
\end{align*}
For every optimizer and every initial learning rate $\alpha_1$, the learning rate after $t$ epochs, $\alpha_t$, is given as
\begin{align*}
    \alpha_t & = 0.95^{\lfloor t / 500 \rfloor} \cdot \alpha_1.
\end{align*}
This is a common schedule in machine learning~\cite{electronics10162029}.
The Adam parameters are chosen following~\cite{deepmind2020jax}: $\beta_1 = 0.9, \beta_2 = 0.999, \epsilon = 10^{-8}, \Bar{\epsilon} = 0$.

We use GELU~\cite{hendrycks2023gaussianerrorlinearunits}, $\sigma(x) = GELU(x) := \frac{x}{2} (1 + \text{erf}\left(\frac{x}{\sqrt{2}} \right))$, a smooth activation because the trunk outputs serve as basis functions for PDE solutions that are smooth for the problems considered here; a piecewise linear activation such as ReLU would restrict the trunk space $\mathcal{T}(N)$ accordingly.
Among smooth activations, GELU performed best in all experiments of the operator learning benchmark~\cite{enyeart2024bestpracticesoperatorlearning}.
\Cref{tab:hyperparams} lists hyperparameters for all figures. \textit{var1}: $\alpha_1 = 10^{-4} \sigma_1^{-2e}$; see~\cref{sec:reweighting}. \textit{var2}, \textit{var3}: varying values for~\cref{fig:spectralbias_branch} (AD: $N=20, w=332$; KdV: $N=50, w=335$; Burgers: $N=20, w=337$), \textit{var4}: $\alpha_1 = 10^{-4}$ for both $T$ and Adam, $\alpha_1 = 4\times10^{-4}$ for $T=\Phi_1 \Sigma_1$ and GD, and $\alpha_1 = 2\times10^{-3}$ for $T=\Phi_1$ and GD.

\begin{longtable}{llrrrrrr}
    \label{tab:hyperparams}               \\
    \toprule
        \textbf{figure}                       & \textbf{example prob.} & \textbf{optimizer} & $\mathbf{\alpha_1}$ & \textbf{depth} & \textbf{width} & $\mathbf{N}$  & \textbf{number of epochs} \\ \midrule
        \ref{fig:graphical_abstract}    & KdV~$\tau=0.2$   & Adam & $2\times10^{-3}$    & $5$            & $100$          &            & $5000$                    \\
        \ref{fig:differentbases}              & KdV~$\tau=0.2$         & Adam               & $2\times10^{-3}$    & $5$            & $100$          &               & $5000$                    \\
        \ref{fig:modeerrors_GD}               & KdV~$\tau=0.2$         & GD                 & $10^{-4}$           & $5$            & $335$          & $50$          & $4000$                    \\
        \ref{fig:modeerrors_weighting_GD}     & KdV~$\tau=0.2$         & GD                 & \textit{var1}       & $5$            & $335$          & $50$          & $4000$                    \\
        \ref{fig:modeerrors_Adam}             & KdV~$\tau=0.2$         & Adam               & $10^{-4}$           & $5$            & $335$          & $50$          & $4000$                    \\
        \ref{fig:sb_synth}                    & synth.~data            & Adam               & $2\times10^{-3}$    & $5$            & $50$           & $5$           & $2000$                    \\
        \ref{fig:spectralbias_branch}         &                        & Adam               & $10^{-4}$           & $5$            & \textit{var2}  & \textit{var3} & $4000$                    \\
        \ref{fig:stacked_vs_unstacked_all}    & KdV~$\tau=0.2$         & Adam               & $10^{-4}$           & $5$            &                & $50$          & $10000$                    \\
        \ref{fig:coupling}                    & KdV~$\tau=0.2$         & GD                 & $10^{-4}$           & $5$            &                & $50$          & $4000$                    \\
        \ref{fig:nosigma_comparison}          & KdV~$\tau=0.2$         &                    & \textit{var4}       & $5$            & $335$          &               & $10000$                   \\
        \ref{fig:2d_errdecomp}                & 2d-heat-eq.            & Adam               & $10^{-4}$           & $10$           & $500$          &               & $10000$                   \\
        \ref{fig:2d_modelosses}               & 2d-heat-eq.            & Adam               & $10^{-4}$           & $10$           & $500$          & $60$          & $10000$                   \\
        \ref{fig:modeerrors_weighting_Adam}   & KdV~$\tau=0.2$         & Adam               & $10^{-4}$           & $5$            & $335$          & $50$          & $4000$                    \\
        \ref{fig:modeerrors_Ada}              & KdV~$\tau=0.2$         &                    & $10^{-4}$           & $5$            & $335$          & $50$          & $4000$                    \\
        \bottomrule
    \caption{\textbf{Hyperparameters of DeepONets and SVDONets used in each figure.}
        A cell is left empty if a figure shows multiple DeepONets and the corresponding hyperparameter values are apparent from the figure.
        The term \textit{var}x denotes a varying value for the different DeepONets which is not apparent from the figure. Each \textit{var}x is explained in the text of this section.
    }
\end{longtable}

\section{Appendix: Input Function Encoding}
\label{app:initcondencoding}

In this work, we restrict input functions to an $L$-dimensional subspace: $p(x) = \sum_{i=1}^L a_i \sin(\omega_i x) \in \operatorname{span}\{\sin(\omega_i x)\}_{i=1}^L \subset \mathcal{C}([0,1])$.
Sampling on a uniform mesh with $M$ interior points $\bar{x}_j = \frac{j}{M+1}$ yields $\hat{p} = \Psi a \in \mathbb{R}^M$, where $\Psi \in \mathbb{R}^{M \times L}$, $\Psi_{ji} = \sin(\omega_i \bar{x}_j)$, and $a \in \mathbb{R}^L$.
Since the input space has dimension $L$, exact reconstruction requires $M \geq L$. If, in addition, $\Psi$ has full column rank, as it does for the sampling grids used in this work, the coefficients can be recovered exactly as $a = \Psi^+ \hat{p}$, yielding zero encoding error.

This restriction is made mainly to simplify the analysis; we can then focus on the contributions of the trunk and branch errors. More generally, the encoding error is bounded from below by the tail $\left(\sum_{k>M} \lambda_k\right)^{1/2}$ of the spectrum $\lambda_1 \geq \lambda_2 \geq \ldots$ of the covariance operator of the input function measure~\cite{Lanthaler_errorestimates_2022}. In our argument, we assume that the sensors are placed sufficiently well for the resulting encoding error to be comparable to this lower bound. Under this assumption, rapidly decaying covariance spectra justify treating the encoding error as negligible, whereas for input distributions with slowly decaying spectra, the encoding error should be retained as a separate term in the decomposition.

\section{Appendix: Testing on Non-Training Coordinates}
\label{app:offgrid}

As discussed, the modified DeepONet's trunk basis is only defined at the $n$ evaluation coordinates $x_i$ of the training grid, while the standard DeepONet's trunk network can be queried at arbitrary coordinates.
Here, we investigate whether either architecture loses accuracy when evaluated at coordinates it was not trained on, using an interpolation of the SVD basis for the modified DeepONet.

We consider the standard example problem (KdV equation, $\tau=0.2$) and refine the evaluation grid by a factor of four: every fourth coordinate of the refined grid coincides with a training coordinate, and the remaining coordinates lie at $1/4$, $1/2$, and $3/4$ of the training grid's cells. The standard DeepONet is evaluated by querying the trunk network directly at the refined coordinates; for the modified DeepONet, each basis vector $t_k = \sigma_k\phi_k$ is interpolated linearly from the training coordinates. The branch networks and their inputs remain unchanged.

\Cref{tab:offgrid} reports the relative test error $\delta_{te}$, computed with respect to the recomputed reference solutions and restricted either to the coordinates shared with the training grid (on-grid) or to the remaining coordinates of the refined grid (off-grid), for the two DeepONets. For each architecture, the on- and off-grid errors are nearly identical. In particular, the linear interpolation of the SVD basis introduces no measurable additional error: the first $N$ basis vectors are well resolved by the training grid, so their interpolation error is far below the networks' approximation error. We expect this to change only when the retained basis vectors are rough relative to the grid spacing, for instance, on significantly coarser grids or for less smooth solutions.

\begin{longtable}{lrr}
    \label{tab:offgrid} \\
    \toprule
        \textbf{DeepONet} & \textbf{on-grid} & \textbf{off-grid} \\
        \midrule
        standard & $2.38\times10^{-1}$ & $2.40\times10^{-1}$ \\
        modified & $6.84\times10^{-2}$ & $6.84\times10^{-2}$ \\
    \bottomrule
    \caption{\textbf{Relative test errors $\delta_{te}$ on- and off-training-grid.} The modified DeepONet's basis vectors are interpolated linearly. Both DeepONets have $N=100$, depth $5$, width $500$, and approximate the KdV equation with $\tau=0.2$.}
\end{longtable}

\section{Appendix: Data}
\label{app:data}

For each example problem, the training and test data sets contain $900$ and $100$ input functions, respectively, both drawn from the same distribution.
The input functions are of the form
\begin{align}
    p(x) = \sum_{i=1}^{L} a_i \sin(\omega_i x), \quad a_i \sim \mathcal{U}([-1, 1]), \nonumber
\end{align}
with problem-specific parameters summarized in~\cref{tab:data}.

\begin{longtable}{lrrrrrrr}
    \label{tab:data}               \\
    \toprule
        \textbf{problem} & $L$ & $\omega_i$ & $\tau$ & \textbf{solver} & $n$ & $M$ & $\text{rank}(A_{tr})$ \\
        \midrule
        AD    & 20 & $2\pi i$ & 0.5, 1.0     & FD + RK~\cite{py-pde}        & 200 & 200 & 20  \\
        KdV   & 5  & $2\pi i$ & 0.2, 0.6, 1.0 & FD + RK~\cite{py-pde}       & 400 & 400 & 400 \\
        Burgers & 5 & $\pi i$ & 0.1, 1.0     & spectral (100 basis) + Euler & 200 & 50  & 100 \\
    \bottomrule
    \caption{\textbf{Data generation parameters for the three example problems.} Here, $L$ is the number of input function modes, $\omega_i$ the corresponding frequencies, and FD + RK denotes the finite-difference method combined with a Runge--Kutta method.
    }
\end{longtable}

The PDEs and boundary conditions for the KdV equation are given in~\cref{sec:data}; the advection-diffusion (AD) and Burgers equations follow the same structure with the differential operators and boundary conditions specified below.

The AD equation is
$\partial_t u + \frac{4}{2\pi} \partial_x u - \frac{0.01}{4\pi^2} \partial_x^2 u = 0$
with periodic boundary conditions on $(0,1)$.
Burgers' equation is
$\partial_t u + u\, \partial_x u - 0.01\, \partial_x^2 u = 0$
with homogeneous Dirichlet boundary conditions on $(0,1)$.

\paragraph{Range of test and training data matrices.}
For the AD equation, the solution operator is linear, so $A_{tr}$ and $A_{te}$ share the same $20$-dimensional range and have rank $20$.
For the KdV equation, $\text{rank}(A_{tr})=400$ equals the spatial dimension $n$, and the standard basis vectors of $\mathbb{R}^{400}$ span $\text{ran}(A_{tr})$, implying $\text{ran}(A_{te}) \subset \text{ran}(A_{tr})$.
For Burgers' equation, the span of the discretized spectral basis functions equals the range of both data matrices.
These containment properties are used in~\cref{app:errordecompose_test_derive}.
Since AD and KdV have $\omega_i = 2 \pi i$, these examples require the input functions to be sampled at $M \geq 2L$ interior meshpoints, in contrast to $M\geq L$ for Burgers.

\section{Appendix: Bound on the Branch Error}
\label{app:branch_error_bound}

As discussed in~\cref{sec:branchtheory} one can derive the following bound on the branch error.
\begin{align*}
    \varepsilon_B &\leq ||T||_2^2 ||B^T - T^+ A||_F^2 = ||T||_2^2 \varepsilon_C =: \varepsilon_D
\end{align*}

\Cref{tab:branch_error_bounds} shows the different error components, and the ratio between $\varepsilon_B$ and $\varepsilon_D$ for different $N$. Recall that, if $\varepsilon_D$ were a tight bound on $\varepsilon_B$, the ratio would be $\approx 1$, i.e., $\log_{10}(\varepsilon_{D,te}/\varepsilon_{B,te}) \approx 0$.
Note that the ratio increases as $N$ increases, since the coefficients of more and more basis functions are poorly approximated.

\begin{longtable}{rlllll}
    \label{tab:branch_error_bounds}                       \\
    \toprule
    $N$ & $\log_{10} \varepsilon_{te}$ & $\log_{10} \varepsilon_{T,te}$ & $\log_{10} \varepsilon_{B,te}$ &  $\log_{10} \varepsilon_{D,te}$ & $\log_{10}(\varepsilon_{D,te}/\varepsilon_{B,te})$ \\ \hline
     2 & 4.31 & 4.31 & 0.43 & 0.61 & 0.18\\
    10 & 3.26 & 3.24 & 2.05 & 3.34 & 1.30\\
    18 & 2.86 & 2.61 & 2.5 & 4.29 & 1.79\\
    50 & 2.59 & 1.62 & 2.54 & 8.29 & 5.75\\
    90 & 2.63 & -0.31 & 2.63 & 13.48 & 10.86\\
    \bottomrule
    \caption{Total error, trunk error, branch error, upper bound on branch error $\varepsilon_D = ||T||_2^2 \varepsilon_C$, and the ratio $\varepsilon_D/\varepsilon_B$.}
\end{longtable}

\section{Appendix: Bases}
\label{app:bases}

We define the bases used in~\cref{sec:trunk}.
Except for the SVD basis vectors, all bases are families of functions $\{ t_k \}_{k=1}^N \subset \mathcal{C}([0,1])$.
For the SVD basis, the $k$-th trunk network output is the $k$-th left singular vector of the training data matrix $A$, i.e., $t_k = \phi_k \in \mathbb{R}^n$.
The learned basis functions are the trunk network outputs of a standard DeepONet.
For the Legendre polynomials, we set $t_k(x) = P_{k-1}(2x-1)$. Here $P_k$ is the $k$-th Legendre polynomial, for instance, $P_0(x) = 1, P_1(x) = x, P_2(x) = (3x^2-1)/2 \ldots$~\cite{AS}.
For the Chebyshev polynomials, we set $t_k(x) = T_{k-1}(2x-1)$. Here $T_k$ is the $k$-th Chebyshev polynomial, for example, $T_0(x) = 1, T_1(x) = x, T_2(x) = 2x^2-1 \ldots$~\cite{AS}.
For the cosine harmonics, we set $t_k(x) = \cos((k-1) \pi x)$.

\section{Appendix: Motivation for DeepONet Modification}
\label{app:trunksigmamotiv}

We choose $T = \Phi_1 \Sigma_1$ for three reasons: (1) $\Phi_1$ gives optimal rank-$N$ approximation ($\varepsilon_T = ||\Sigma_2||_F^2$); (2) using $\Sigma_1$ makes branch targets semi-orthogonal ($||v_i||_2^2 = 1$), a common normalization in machine learning~\cite{bishop, hastie, hu-revisit, huang-normal}; (3) simplifies analysis with straightforward unweighted mode loss $L_i$.

\section{Appendix: Deriving the Mode-Based Error Decomposition for Test Data}
\label{app:errordecompose_test_derive}

We can write $A_{te} = \Phi \Sigma W^T + A_{te}^\perp$, where $W = (\Sigma^+ \Phi^T A_{te})^T$ and $A_{tr}^T A_{te}^\perp = 0$. Furthermore, we can partition $W = [W_1~W_2]$ with $W_1 \in \mathbb{R}^{m_{te} \times N}$. Thus, $W_1$ contains the first $N$ optimal coefficients for the test data.
Then, we can decompose the test error as
\begin{align*}
    \varepsilon_{te}  &= || \Phi_1 \Sigma_1 B_{te}^T - \Phi_1 \Sigma_1 W_1^T - \Phi_2 \Sigma_2 W_2^T - A_{te}^\perp ||_F^2 = || \Sigma_1 B_{te}^T -  \Sigma_1 W_1^T ||_F^2 + ||\Phi_2 \Sigma_2 W_2^T + A_{te}^\perp||_F^2 \\
     &= \sum_{i=1}^N \sigma_i^2 || b_{i, te} -  w_i ||_2^2 + \underbrace{||\Phi_2 \Sigma_2 W_2^T+A_{te}^\perp||_F^2}_{=: \varepsilon_{T,te}}.
\end{align*}
In our case we have $A_{te}^\perp = 0$, since $\text{ran}(A_{te}) \subset \text{ran}(A_{tr})$; see~\cref{app:data}.

Note: $A_{te} = \Phi \Sigma W^T$ is not an SVD, so $\Phi_1$ lacks optimality for test data. Unlike training ($V_1, V_2$ semi-orthogonal), $W_1, W_2$ are not semi-orthogonal, so test trunk error could exceed training trunk error. However, this does not occur for our examples; see~\cref{tab:trunk_train_test_errors}.

\begin{longtable}{llrrrrrrr}
    \label{tab:trunk_train_test_errors}                     \\
    \toprule
        \multicolumn{2}{l}{\textbf{problem}} & $\log_{10}\delta_{T, tr}$ & $\log_{10}\delta_{T, te}$ & $\log_{10}\delta_{T, tr}$ & $\log_{10}\delta_{T, te}$ & $\log_{10}\delta_{T, tr}$ & $\log_{10}\delta_{T, te}$          \\
        \midrule
                                             &                                    & $N=10$                             & $N=10$                             & $N=15$                             & $N=15$                             & $N=20$                             & $N=20$ \\
        \midrule
        AD                                   & $\tau=0.5$                         & -0.41                              & -0.41                              & -0.78                              & -0.78                              & -12.6                              & -12.6  \\
        AD                                   & $\tau=1.0$                         & -0.69                              & -0.69                              & -1.32                              & -1.32                              & -13.1                              & -13.1  \\
        \midrule
                                             &                                    & $N=30$                             & $N=30$                             & $N=50$                             & $N=50$                             & $N=70$                             & $N=70$ \\
        \midrule
        KdV                                  & $\tau=0.2$                         & -1.48                              & -1.46                              & -2.19                              & -2.10                              & -2.94                              & -2.84  \\
        KdV                                  & $\tau=0.6$                         & -1.04                              & -1.03                              & -1.90                              & -1.88                              & -2.71                              & -2.59  \\
        KdV                                  & $\tau=1.0$                         & -1.00                              & -0.99                              & -1.91                              & -1.93                              & -2.70                              & -2.70  \\
        Burgers                              & $\tau=0.1$                         & -1.36                              & -1.29                              & -1.90                              & -1.75                              & -2.42                              & -2.20  \\
        Burgers                              & $\tau=1.0$                         & -3.56                              & -3.35                              & -5.53                              & -5.10                              & -7.54                              & -6.75  \\
        \bottomrule
    \caption{Relative training and test trunk errors $\delta_T$ for all example problems (advection-diffusion (AD), Korteweg-de Vries (KdV) and Burgers) and various $N$.}
\end{longtable}

Furthermore, since the trunk error is determined by the choice of $N$, it can be omitted from the loss used in the neural network training.
Thus, we use
\begin{align*}
    \mathcal{L}_{tr} \coloneqq \frac{1}{n_{tr} m_{tr}}|| \Sigma_1 B_{tr}^T - \Sigma_1 V_1^T ||_F^2 = \frac{1}{n_{tr} m_{tr}} \varepsilon_{B,tr}, \qquad
    \mathcal{L}_{te} \coloneqq \frac{1}{n_{te} m_{te}}|| \Sigma_1 B_{te}^T - \Sigma_1 W_1^T ||_F^2 = \frac{1}{n_{te} m_{te}} \varepsilon_{B,te}, \nonumber
\end{align*}
as training and test loss for the modified DeepONet.
This corresponds to only measuring the branch error, not the full approximation error of the matrix $A$.
This implies
\begin{align*}
    \mathcal{L}_{tr} = \frac{1}{n m_{tr}} \sum_{i=1}^N \sigma_i^2 L_{i,tr}, \qquad
    \mathcal{L}_{te} = \frac{1}{n m_{te}} \sum_{i=1}^N \sigma_i^2 l_i.
\end{align*}
To facilitate the comparison between the unweighted mode losses for test and training data, we define the unweighted test loss of mode $i$ as $L_{i,te} = \frac{m_{tr}}{m_{te}} l_i = \frac{m_{tr}}{m_{te}} ||b_{i,te}-w_i||_2^2$. This implies
\begin{align*}
    \mathcal{L}_{te} & = \frac{1}{n m_{tr}} \sum_{i=1}^N \sigma_i^2 L_{i,te}.
\end{align*}
Thus, the unweighted test base loss is $\frac{m_{tr}}{m_{te}} ||w_i||_2^2$.

\section{Appendix: Comparative Mode Loss Analysis for Different Trunks}
\label{app:trunk_phi_vs_normal}

Since the impact of choosing $T=\Phi_1 \Sigma_1$ over a different $T=\Phi_1 C$, introduced in~\cref{sec:trunktheory}, is hard to predict analytically, it is unclear a priori whether the analysis with, e.g., $T=\Phi_1$ will differ from that with $T=\Phi_1 \Sigma_1$. As the following experiments show, the analysis is at least qualitatively similar for both cases.

Firstly, we can derive an error decomposition for $T=\Phi_1$. Here, we derive it for the training loss. However, a similar decomposition can be derived for the test data.
\begin{align*}
    \varepsilon_{tr} & = ||\tilde{A}_{tr}-A_{tr}||_F^2 = || \Phi_1 B_{tr}^T - \Phi_1 \Sigma_1 V_1^T - \Phi_2 \Sigma_2 V_2^T ||_F^2 = || \Phi_1 B_{tr}^T - \Phi_1 \Sigma_1 V_1^T||_F^2 + || \Phi_2 \Sigma_2 V_2^T ||_F^2 \nonumber \\
    & = || B_{tr}^T - \Sigma_1 V_1^T ||_F^2 + ||\Sigma_2||_F^2 =  \sum_{j=1}^N \underbrace{|| b_{j, tr}-\sigma_j v_j ||_2^2}_{=: \tilde{L}_{j, tr}} + ||\Sigma_2||_F^2
\end{align*}

\begin{figure}
    \centering
    \includegraphics[width=1.0\textwidth]{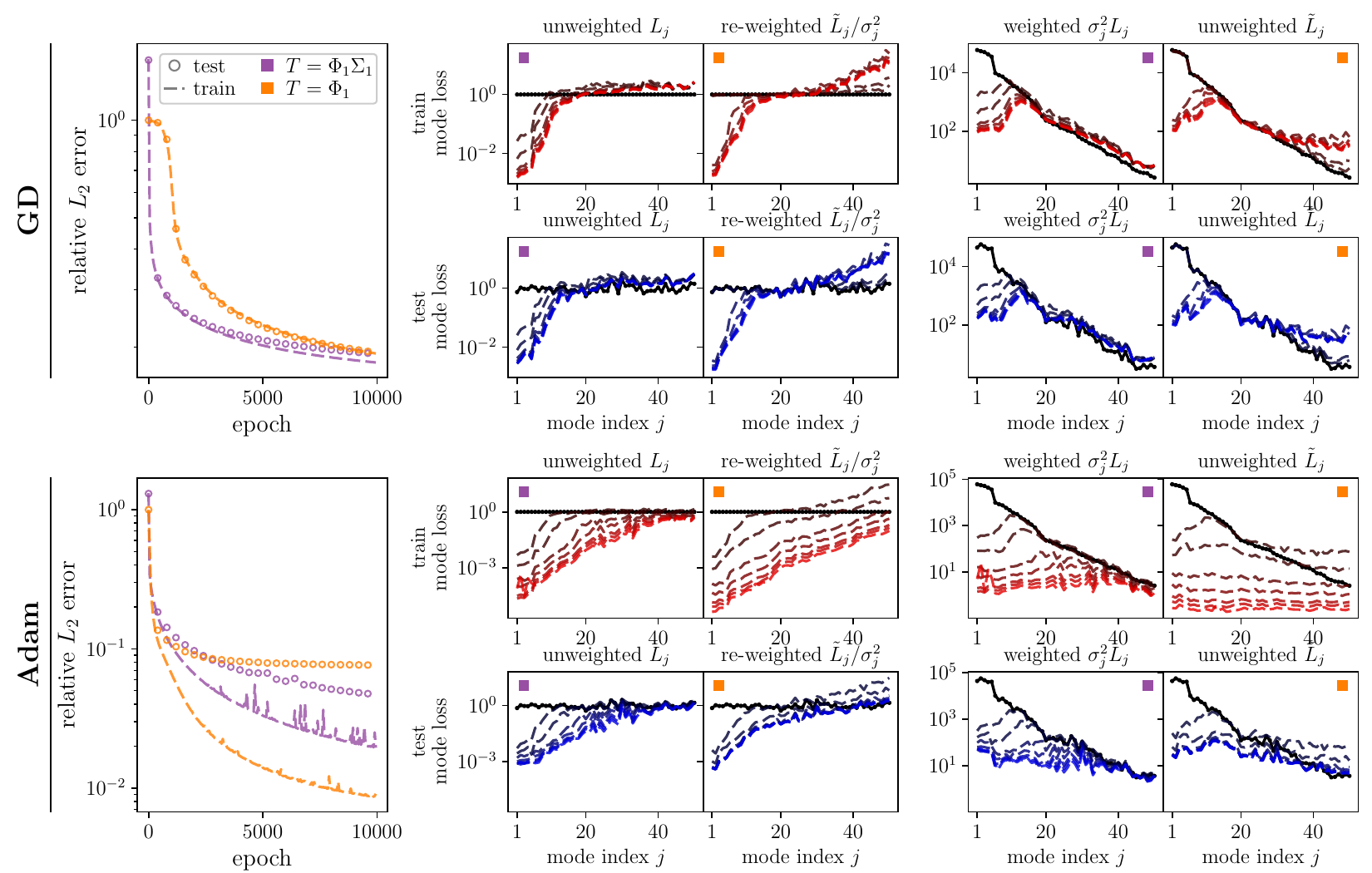}
    \caption{
        \textbf{Comparison of modified DeepONets with $T=\Phi_1$ and $T=\Phi_1\Sigma_1$, trained with GD and Adam.}
        \textbf{Top / bottom block:} DeepONets trained with GD / Adam.
        \textbf{Left panel of each block:} Relative error $\delta = ||A - \Tilde{A}||_F/||A||_F$ for training (dashed lines) and test (circles) data over $\num{10000}$ epochs, for $T=\Phi_1\Sigma_1$ (purple) and $T=\Phi_1$ (orange).
        \textbf{Remaining panels:} Training (upper row) and test (lower row) mode losses at different training stages, colored from gray (initial) to red/blue (final), with the respective base losses in black.
        Panels marked with a purple square show the DeepONet with $T=\Phi_1\Sigma_1$, those marked with an orange square the one with $T=\Phi_1$.
                The two center columns show the mode losses relative to the $\sigma_j^2$, while the two right columns show the absolute contributions of mode $j$ to the error; the panels within each pair are thus directly comparable. The curves correspond to only one DeepONet; no averaging over multiple seeds was done.
    }
    \label{fig:nosigma_comparison}
\end{figure}

Here, we term $\tilde{L}_j$ the unweighted mode losses and define $\tilde{L}_{j, tr}/\sigma_j^2$ as the re-weighted mode losses. Note that $\tilde{L}_j$ and $\sigma_j^2 L_j$ both measure the absolute contribution of mode $j$ to $\varepsilon_{tr}$, whereas $\tilde{L}_j/\sigma_j^2$ and $L_j$
measure it relative to $\sigma_j^2$. This motivates comparing the two choices of $T$ crosswise: $\tilde{L}_j$ against $\sigma_j^2 L_j$, and $\tilde{L}_j/\sigma_j^2$ against $L_j$.
\Cref{fig:nosigma_comparison} shows the relative errors and mode losses for DeepONets trained with Adam and GD, for both choices of $T$. In all four settings, the large modes are learned first and best.
However, there are two noteworthy differences.
First, with Adam and $T=\Phi_1$, the re-weighted training mode losses $\tilde{L}_{j,tr}/\sigma_j^2$ are eventually driven below the base loss for every $j$, whereas for $T=\Phi_1\Sigma_1$ the losses of the small modes stagnate at the base loss. This improved training fit does not transfer: the corresponding test mode losses exceed the base loss for large $j$, and the test error is higher than for $T=\Phi_1\Sigma_1$ despite the lower training error. This matches our general observation of high overfitting in small modes; see~\cref{sec:modelossdistr}.
        Second, with GD and $T=\Phi_1$, the small modes are not even reduced to the base loss, with $\tilde{L}_{j,tr}/\sigma_j^2$ rising above it for large $j$, while for $T=\Phi_1\Sigma_1$ they stay at the base loss throughout training. However, this has a negligible effect on the total loss.
Finally, note that in the GD case the learning rate determines which of the two choices of $T$ attains the lower relative $L_2$ error; since our aim is analysis rather than performance, we selected learning rates giving comparable final errors.

        On a higher level, this comparison shows that the choice of $T$ affects the mode loss distribution. However, some observations appear to hold generally. In particular, DeepONets attain their lowest unweighted losses $L_j$, or re-weighted losses $\tilde{L}_j/\sigma_j^2$, for the largest modes; this can be understood as a consequence of gradient pressure.
However, one can conjure examples (such as our synthetic data with $\alpha=-0.2, \beta=-0.01$,~\cref{sec:spectralbias}) where spectral bias counteracts the gradient pressure.
        Thus, we do not believe that our specific choice of $T$ is particularly important; rather, $T=\Phi_1\Sigma_1$ is one choice that yields an elegant decomposition (\cref{eq:modelossdecompose_train}), desired normalization properties (\cref{app:trunksigmamotiv}), and the general, largely $T$-independent mode loss distribution.

\section{Appendix: 2D Problems}
\label{app:2d}

All example problems in the main text are posed on one-dimensional spatial domains, and their input functions are drawn from sine series with uniformly distributed random coefficients; see~\cref{app:data}.
In this appendix, we test whether our main observations (the distribution of the error between trunk and branch in~\cref{sec:trunk} and the characteristic mode loss distribution in~\cref{sec:branch}) carry over to a two-dimensional problem and to input functions drawn from a different distribution, namely Gaussian random fields (GRFs).

\subsection{Problem and Data}
\label{app:2d_data}

We consider the heat equation on the unit square $\Omega = (0,1)^2$ with homogeneous Dirichlet boundary conditions,
\begin{alignat*}{2}
    \frac{\partial u}{\partial t} &= D \, \Delta u &\qquad& \text{in } \Omega \times (0, \tau],\\
    u &= 0 &\qquad& \text{on } \partial\Omega \times (0, \tau],\\
    u(\cdot, 0) &= p(\cdot) &\qquad& \text{in } \Omega,
\end{alignat*}
with diffusion coefficient $D=1$ and evolution time $\tau = 0.004$.
The eigenfunctions of the (negative) Dirichlet Laplacian on $\Omega$ are $s_{ij}(x,y) = \sin(i \pi x) \sin(j \pi y)$ with eigenvalues $\lambda_{ij} = (i^2+j^2)\pi^2$ for $i,j \geq 1$, and the solution operator is diagonal in this basis: a mode with amplitude $c_{ij}$ evolves exactly as $c_{ij}(\tau) = c_{ij}(0) \, e^{-D \lambda_{ij} \tau}$.
We exploit this to compute the target solutions spectrally (via the discrete sine transform), so the data contain no time-integration error; the only discretization is spatial sampling.

We consider two families of input functions:
\begin{enumerate}
    \item[(a)] \textit{Tensor-product sine series}, the two-dimensional analog of the input functions in~\cref{app:data}:
    \begin{align*}
        p(x,y) = \left( \sum_{i=1}^{K} a_i \sin(i \pi x) \right) \left( \sum_{j=1}^{K} b_j \sin(j \pi y) \right), \qquad a_i, b_j \sim \mathcal{U}([-1,1]),
    \end{align*}
    with $K = 8$ modes per axis. These input functions lie in the $K^2 = 64$-dimensional span of the eigenfunctions $\{ s_{ij} \}_{i,j=1}^{K}$; consequently, the solutions lie in the span of the same $64$ (evolved) eigenfunctions and $\text{rank}(A_{tr}) = 64$.
    \item[(b)] \textit{Boundary-compatible GRFs.} GRFs are a common choice of input distribution in the operator learning literature~\cite{Bhattacharya, deeponet, li_fno}; however, a stationary GRF, e.g., with the squared-exponential (RBF) covariance kernel~\cite{gpml}, does not vanish on $\partial\Omega$ and is thus not an admissible initial condition for the homogeneous Dirichlet problem. We therefore synthesize the fields directly in the Dirichlet--Laplacian eigenbasis:
    \begin{align*}
        p(x,y) = \sum_{i,j \geq 1} c_{ij} \, s_{ij}(x,y), \qquad c_{ij} \sim \mathcal{N}\!\left(0, \, e^{-\ell^2 \lambda_{ij}/2}\right) \text{ independently}.
    \end{align*}
    This is the Karhunen--Lo\`eve expansion~\cite{lord_spde_book} of a mean-zero Gaussian field whose covariance operator is a function of the Dirichlet Laplacian~\cite{lindgren_spde}: the mode variances follow the power spectral density $\exp(-\ell^2 ||\mathbf{k}||^2/2)$ of the RBF kernel~\cite{gpml} with length scale $\ell$, with the Laplacian eigenvalue $\lambda_{ij}$ taking the role of the squared wavenumber $||\mathbf{k}||^2$.
    Hence, $\ell$ controls the smoothness of the samples in the same way as for a stationary RBF-kernel GRF, while every sample satisfies the boundary conditions exactly; the price is that the variance is suppressed near $\partial\Omega$.
    We generate one dataset with $\ell = 0.05$.
\end{enumerate}
        In contrast to family (a) and to all 1D examples, the input functions of family (b) are not confined to a low-dimensional subspace, and their coefficients are Gaussian and not uniform; they thus constitute a genuinely different input distribution. \Cref{tab:data2d} contains the ranks of the input function matrices; for both datasets, these ranks are significantly larger than the ranks $L \in \{5,20\}$ of the 1D problems.

All functions are sampled on the $64 \times 64$ interior grid, i.e., $n = M = 4096$.
Since both input function families are (grid-)band-limited in the sine basis, $p$ is exactly reconstructible from $\hat{p}$, so there is no encoding error, analogously to~\cref{app:initcondencoding}.
Each dataset contains $m = \num{5000}$ input functions, split into $m_{tr} = \num{4500}$ training and $m_{te} = 500$ test samples.
\Cref{tab:data2d} summarizes the datasets.
        Note that, due to the fast decay of the evolution factors $e^{-D \lambda_{ij} \tau}$, the training data matrix of the GRF dataset has a numerical rank far below $\min(n, m_{tr})$.
\Cref{fig:2d_errdecomp} (left) shows the normalized singular values of the training data matrices for some 1D datasets and the 2D datasets showing the significantly slower singular value decay in the 2D datasets.

\begin{longtable}{lrrrrrrr}
    \label{tab:data2d} \\
    \toprule
    \textbf{dataset} & \textbf{family} & \textbf{parameters} & $\tau$ & $n = M$ & $m_{tr}$ / $m_{te}$ & $\text{rank}(A_{tr})$ & $\text{rank}(P)$ \\
    \midrule
    sine        & (a) & $K=8$                    & 0.004 & 4096 & 4500 / 500 & 64    & 64        \\
    GRF  & (b) & $\ell = 0.05$            & 0.004 & 4096 & 4500 / 500 & $\approx 300$ & $\approx 2300$\\
    \bottomrule
    \caption{\textbf{Data generation parameters for the two-dimensional heat equation datasets.} All datasets share the exact spectral solver. We report the rank for both the training data matrix $A_{tr}$ and the matrix containing the input vectors $P \in \mathbb{R}^{M \times m_{tr}}$. For the sine datasets, these ranks are exact; for the GRF datasets, we report the numerical rank (singular values above $10^{-8} \sigma_1$).}
\end{longtable}

Regarding the cost of the SVD in higher spatial dimensions: for a grid with $k$ points per dimension in $d$ dimensions, the data matrix has $n = k^d$ rows and $m$ columns, and a dense SVD costs $\mathcal{O}(\min(n,m)^2 \max(n,m))$ operations. However, only the leading $N \ll \min(n,m)$ modes are required for the trunk basis, which truncated or randomized SVD algorithms compute in $\mathcal{O}(nmN)$ operations~\cite{halko}. This is linear in $n = k^d$, whereas storing the data matrix is also linear in $n$ and generating it costs at least that (one PDE solve per column, each at least linear in the number of grid points). The exponential growth of the grid with $d$ thus affects the
factorization at most as severely as the training data itself; see~\cref{sec:discussion}. For the datasets in this appendix ($n = 4096$, $m_{tr} = 4500$), even the dense SVD takes only $\approx15$ seconds on an AMD EPYC 7662.

\subsection{Trunk and Branch Errors}
\label{app:2d_trunk_branch}

        \begin{figure}[!ht]
    \centering
    \includegraphics[width=0.95\textwidth]{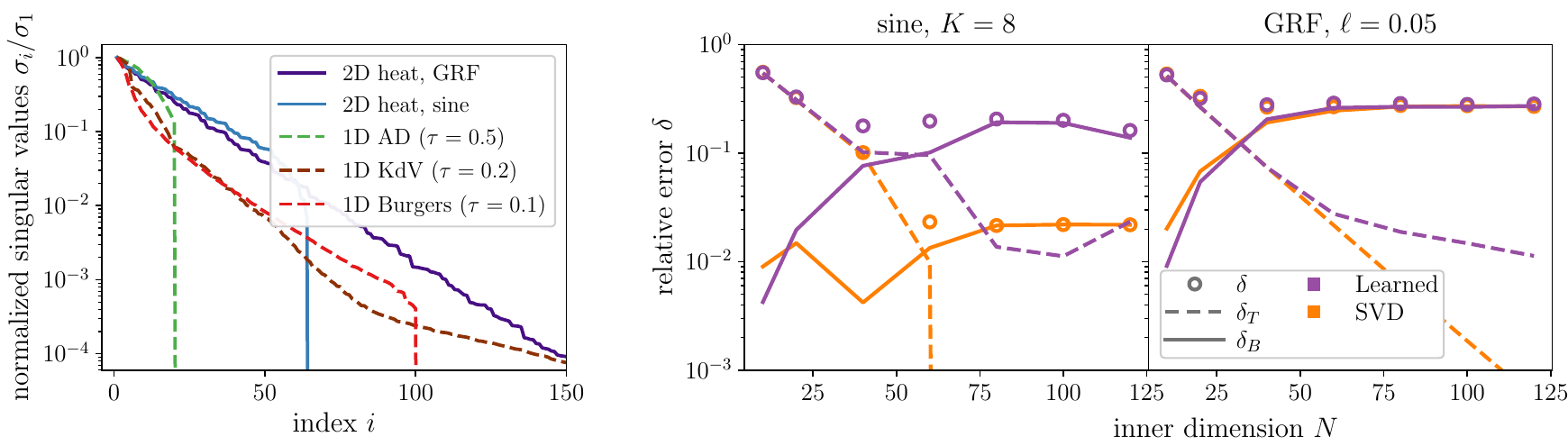}
    \caption{
    \textbf{Left: Singular values of 1D and 2D datasets.}
    \textbf{Center/right: Relative total, trunk and branch errors for the 2D heat equation plotted over $N$.}
    The two panels show the errors for sine input functions ($K=8$, center) and GRFs ($\ell = 0.05$, right).
    \textbf{Colors:} Purple: learned trunk (standard DeepONet), orange: SVD basis.
    \textbf{Line-symbols:} Total error $\delta$ (circle), trunk error $\delta_T$ (dashed), and branch error $\delta_B$ (solid); see~\cref{fig:differentbases}.
    For the sine dataset, the SVD trunk error vanishes for $N \geq \text{rank}(A_{tr}) = 64$.
                Each data point corresponds to one DeepONet; no averaging over multiple seeds was done.
    }
    \label{fig:2d_errdecomp}
\end{figure}

\Cref{fig:2d_errdecomp} (center and right) shows the relative total, trunk, and branch errors over the inner dimension $N$ for the two datasets, for both the learned and the SVD trunk basis.
The picture from~\cref{sec:trunk}, illustrated in~\cref{fig:differentbases}, carries over qualitatively to both datasets:
the trunk error $\delta_T$ decreases with $N$, while the branch error $\delta_B$ increases with $N$ and then plateaus.
Consequently, for large $N$, the branch error dominates and the total error $\delta$ stagnates despite the decreasing trunk error.
For the sine dataset, the SVD trunk error is exactly zero for $N \geq 64$ by construction, which makes the identity $\delta = \delta_B$ exact there; the total error nevertheless plateaus, isolating the branch as the sole error source.

Note that for the sine dataset, the modified DeepONet with the SVD basis achieves a noticeably lower total error than the standard DeepONet, whereas both perform nearly identically for $\ell = 0.05$.
However, this does not affect the main observation: in two dimensions as well, the branch error, not the trunk error, limits the DeepONet's accuracy for large $N$.

\subsection{Mode Losses}
\label{app:2d_modes}

        \begin{figure}[!ht]
    \centering
    \includegraphics[width=0.95\textwidth]{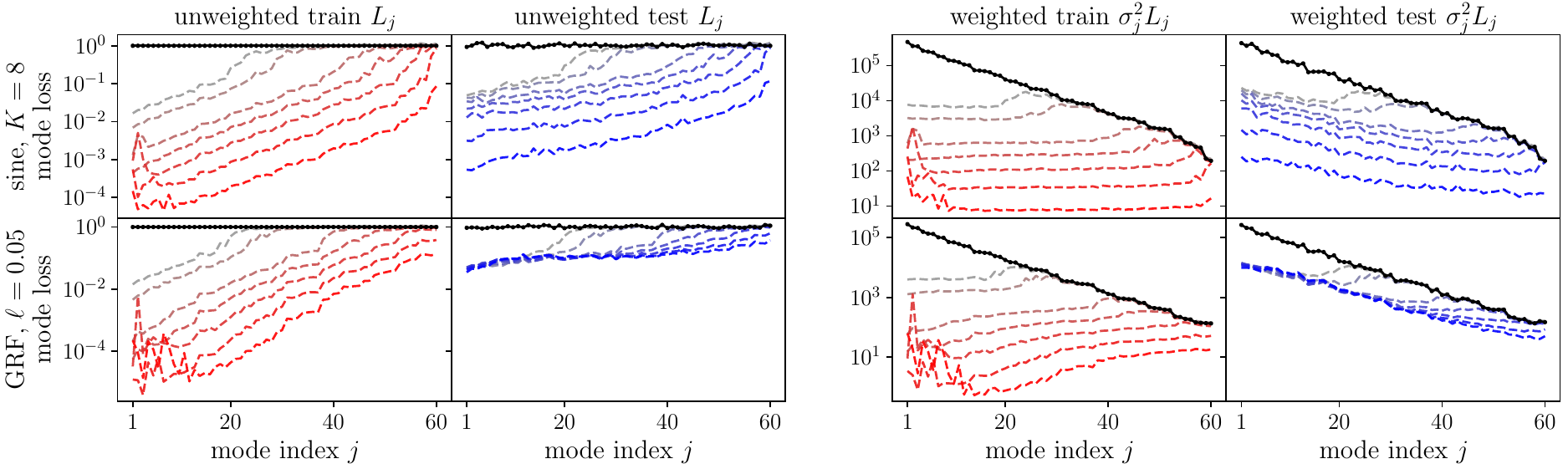}
    \caption{\textbf{Mode losses for the 2D heat equation.}
        Unweighted (left) and weighted (right) training (red) and test (blue) mode losses at different training stages, colored from gray (initial) to red/blue (final), shown as dashed lines.
        The respective base losses are shown as black dot-solid lines.
        \textbf{Rows:} sine input functions ($K=8$) and GRFs with $\ell = 0.05$.
                The curves correspond to only one DeepONet; no averaging over multiple seeds was done.
    }
    \label{fig:2d_modelosses}
\end{figure}

\Cref{fig:2d_modelosses} shows the unweighted and weighted mode losses of modified DeepONets with $N=60$ over the course of training, analogously to~\cref{fig:modeerrors_Adam}.
The characteristic mode loss distribution observed for the 1D examples emerges for both datasets: the unweighted mode losses increase with the mode index; the large modes are learned first and most accurately.
        However, due to the slower singular value decay in the 2D datasets, the (weighted) test loss remains dominated by the large modes, not the intermediate modes.
The GRF dataset additionally exhibits very strong overfitting: after a few hundred epochs, the test mode losses stagnate at an approximately uniform level across all modes, while the training losses continue to decrease by several orders of magnitude.

        Overall, we observe similar patterns in the mode loss distributions of the 2D and 1D problems.
This suggests that the mode loss distribution and its driving mechanism, specifically the singular-value weighting of the gradients partially compensated by Adam, are not artifacts of the 1D problems or of the uniform sine-series input distribution.
\section{Appendix: Further Optimization Algorithms}

\subsection{Re-Weighting with Adam}
\label{app:adam_reweight}

\begin{figure}[!ht]
    \centering
    \includegraphics{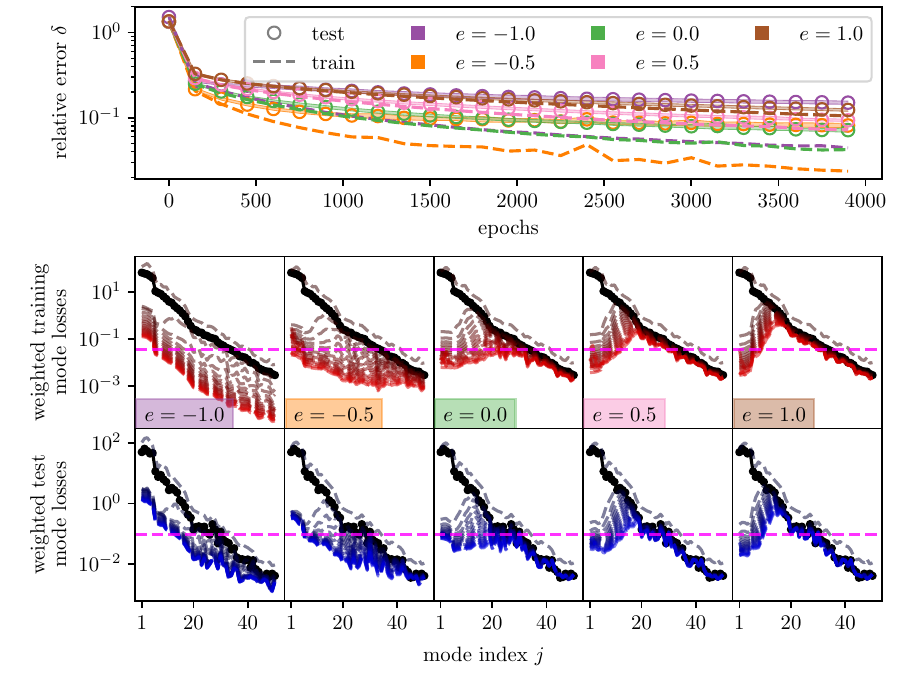}
    \caption{
        \textbf{Model performance across different exponents $e$ and training epochs for modified DeepONets trained using Adam.}
        Layout and parameters identical to~\cref{fig:modeerrors_weighting_GD}, with Adam replacing GD.
        \textbf{Top panel:} Relative error $\delta = ||A- \Tilde{A}||_F/||A||_F$ for both training (dashed lines) and test (circles) data across different exponents ($e = -1.0, -0.5, 0.0, 0.5, 1.0$) over $4000$ epochs.
        \textbf{Center and bottom row:}
        Weighted training (center row) and test (bottom row) mode losses at different training steps, colored from gray (initial) to red/blue (final).
        Each column corresponds to a different exponent $e$.
        The third column shows the DeepONet trained using the standard loss ($e=0$).
        The center and bottom row plots also contain the respective base losses in black, and a pink dashed horizontal line marking the maximum mode loss in the last training epoch of $e=0$, facilitating comparison between different exponents $e$.
        For visual clarity, the training-loss curve in the top panel and the plots in the center and bottom rows have no shaded bands.
    }
    \label{fig:modeerrors_weighting_Adam}
\end{figure}

\Cref{fig:modeerrors_weighting_Adam} shows the relative error and the weighted mode losses for DeepONets trained using Adam with different exponents $e$.

\subsection{AdaGrad}
\label{app:adagrad}

\begin{figure}[!ht]
    \centering
    \includegraphics{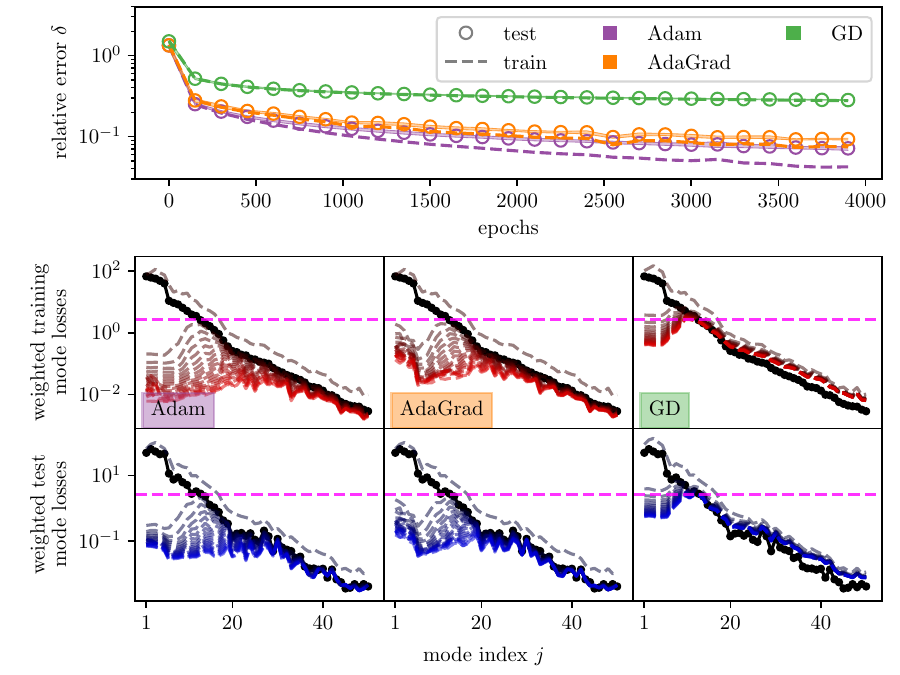}
    \caption{
        \textbf{Model performance for DeepONets trained with different optimizers.}
        \textbf{Top panel:} Relative error $\delta = ||A- \Tilde{A}||_F/||A||_F$ for both training (dashed lines) and test (circles) data for different optimizers (purple: Adam, orange: AdaGrad, green: GD) over $4000$ epochs.
        \textbf{Center and bottom row:}
        Weighted training (center row) and test (bottom row) mode losses at different training steps, colored from gray (initial) to red/blue (final).
        The center and bottom row plots also contain the respective base losses in black, and a pink dashed horizontal line marking the maximum mode loss in the last training epoch of the DeepONet trained with GD, facilitating comparison between different algorithms.
        For visual clarity, the training-loss curve in the top panel and the plots in the center and bottom rows have no shaded bands.
    }
    \label{fig:modeerrors_Ada}
\end{figure}

\Cref{fig:modeerrors_Ada} compares the weighted mode losses for DeepONets trained with GD, Adam, and AdaGrad. We observe that the overall and mode-specific test losses of the DeepONets trained with Adam and AdaGrad are very similar.

\section{Appendix: Spectral Bias}
\label{app:spectralbias}

In this section, we provide more information on our spectral bias analysis.
The generation of the synthetic data, used in~\cref{sec:spectralbias} to analyze spectral bias in the branch network for data with curated spectral properties, is described in~\cref{app:syntheticdata}.
The methods for estimating the frequency of the right singular functions for the PDE-derived data are described in~\cref{app:frequencyestimation}.
In~\cref{sec:fourier_closing} these methods are evaluated on the synthetic data.

\subsection{Synthetic Data for Spectral Bias}
\label{app:syntheticdata}

As discussed, the branch network approximates the right singular vectors $v_i$ of $A$. Thus, to study the effect of (a) singular values and (b) spectral bias in the branch network, we vary the singular values and generate the right singular vectors with the desired frequencies: $A = \Phi \Sigma V^T$. The left singular matrix can thus be chosen as any orthogonal matrix $\Phi$. For the singular values, we choose $\sigma_j = \exp(\beta j)$. We now discuss how to manufacture $V$ with the desired frequencies.
A prototypical multivariate function $g_k : \mathbb{R}^M \to \mathbb{R}$ with mean frequency $f = \|k\|_2$ is $g_k(p) = \sin(2\pi k^T p)$.
We would ideally set the right singular functions to $\rho_j = g_{f_j z}$ for a family of frequencies $\{f_j\}_{j=1}^N$ and a unit vector $z \in \mathbb{R}^M$, but the resulting vectors are not necessarily orthogonal.
We address this by (1) choosing $p \in \mathbb{R}^M$ and direction vectors such that the resulting vectors are approximately orthogonal, and (2) applying Gram--Schmidt orthogonalization.

\Cref{alg:generate_rsf} contains the pseudocode.
The algorithm generates $m$ sample points concentrated near the $\mathbf{1}_M = (1~\ldots~1)^T$ direction, then constructs a sinusoidal candidate $\sin(2\pi f_j d^T x)$ for each frequency $f_j$ using randomly sampled directions $d \sim \mathcal{N}(\mathbf{1}_M, I_M)$.
Candidates are accepted only if their inner product with all previously selected functions is below a threshold of $t = 0.05$, with up to $K$ attempts per frequency.
A final Gram--Schmidt step enforces orthogonality of the basis $V = [v_1, \ldots, v_N]$.
Since we are not aware of a reliable method to quantify the frequency of a multivariate function, as discussed in~\cref{app:frequencyestimation}, we estimate the mean frequency of $\rho_j$ to be $f_j$.
\begin{algorithm}
    \caption{Generation of right singular functions with prescribed frequencies}
    \label{alg:generate_rsf}
    \begin{algorithmic}[1]
        \Require Number of trials $K$, desired frequencies $F = \{ f_1, \dots, f_N \}$, number of samples $m$, input dimension $M$
        \Statex

        \Comment{Generate sample points in $\mathbb{R}^M$}
        \For{$i = 1$ to $m$}
        \State Draw $a \sim \mathcal{U}([0,1])$ and $b \sim \mathcal{U}([-1,1]^M)$
        \State $x_i \gets a \, (\mathbf{1}_M + 0.1\, b)$
        \EndFor
        \State $X \gets \{ x_i \}_{i=1}^m$
        \State Set inner-product threshold $t \gets 0.05$
        \Statex

        \Comment{Construct frequency-specific orthogonal functions}
        \For{$j = 1$ to $N$}
        \For{$k = 1$ to $K$}
        \State Draw direction $d \sim \mathcal{N}(\mathbf{1}_M, I_M)$
        \State Define $w_i \gets \sin(2\pi f_j \, d^\top x_i)$ for all $i = 1, \dots, m$
        \State $w \gets [w_1, \dots, w_m]^\top$
        \If{$\max_{1 \leq a < j} | w^\top v_a | < t$}
        \State $v_j \gets w / \|w\|_2$
        \State \textbf{break}
        \EndIf
        \EndFor
        \EndFor
        \Statex

        \Comment{Final orthogonalization step}
        \State Orthogonalize $\{ v_1, v_2, \dots, v_N \}$ using Gram--Schmidt.
        \State \Return $V = [v_1, v_2, \dots, v_N]$
    \end{algorithmic}
\end{algorithm}

\subsection{Frequency Estimation for Multivariate Functions}
\label{app:frequencyestimation}

We seek to characterize the spectrum of $\rho_k : \mathbb{R}^M \to \mathbb{R}$ from the samples $\rho_k(p_j) = V_{jk}$.
Since $\rho_k$ has multi-dimensional input sampled at irregular locations, computing its Fourier transform directly is infeasible~\cite{xu-sb2}.
We employ three methods: the total variation (TV) norm~\cite{RUDIN1992259, getreuer}, the Laplacian energy (LE)~\cite{spielman}, and Fourier transforms of projected data~\cite{xu-sb2}, described in~\cref{sec:TVnorm,sec:LaplacianEnergy,sec:proj_fourier}.
Both the TV norm and LE rely on a $k$-nearest neighbors algorithm ($k$ denotes the number of neighbors in this context).
For ease of notation, we call the outputs of all methods frequencies, although the TV norm and LE technically compute related quantities.

To compare these methods, we use the relative frequency $f_i / (\max_j f_j)$.
Since we are primarily interested in qualitative trends, specifically whether $f(\rho_k)$ increases or decreases with $k$, rather than exact values, this comparison is meaningful, as confirmed empirically in~\cref{sec:fourier_closing}.
In the following sections, we discuss frequency estimation methods for general functions $g : D \subset \mathbb{R}^d \to \mathbb{R}$ observed through samples $\{x_j, y_j\}_{j=1}^m$ with $y_j = g(x_j)$. Thus, for $g = \rho_k$, we have $x_j = p_j$ and $y_j = V_{jk}$.

\subsubsection{Total Variation Norm}
\label{sec:TVnorm}

The total variation (TV) norm of a differentiable function $g : D \subset \mathbb{R}^d \to \mathbb{R}$ is
$\|g\|_{TV} = \int_D \|\nabla g\|_2 \, dx$.
We approximate it using $k$-nearest neighbors~\cite{kNN}:
\begin{align*}
    \|g\|_{TV} \approx \frac{1}{mk} \sum_{i=1}^m \sum_{j \in N_k(i)} \frac{|y_i - y_j|}{\|x_i - x_j\|_2},
\end{align*}
where $N_k(i)$ denotes the $k$ nearest neighbors of $x_i$.
This estimates the gradient norm via finite difference quotients between neighboring samples.
We use $f(g) = \|g\|_{TV}$ as a proxy for the frequency.

\subsubsection{Laplacian Energy}
\label{sec:LaplacianEnergy}

Consider a weighted graph $G=(V,E,w)$ with Laplacian matrix $L \in \mathbb{R}^{v \times v}$ defined by
\begin{align*}
    L_{ij} =
    \begin{cases}
        -w((i,j))                  & \text{if } (i,j) \in E, \\
        \sum_{l \in N(i)} w((i,l)) & \text{if } i=j, \\
        0                          & \text{else,}
    \end{cases}
\end{align*}
where $N(i)$ is the set of adjacent vertices~\cite{spielman}.
For undirected graphs, $y^T L y = \sum_{(i,j) \in E} w((i,j)) (y_i - y_j)^2 \geq 0$, and the eigenpairs $(\mu_i, u_i)$ of $L$ correspond to increasing frequencies with index $i$.

To apply this to a function $g$ and the samples $\{ x_i, y_i \}$, we construct a graph with vertices corresponding to samples and edges from a $k$-nearest neighbors algorithm.
Edge weights are computed via a Gaussian kernel:
\begin{align*}
    w((i,j)) = \exp\left( -\frac{k}{2} \frac{\|x_i-x_j\|_2^2}{\sum_{l \in N(i)} \|x_i-x_l\|_2^2} \right).
\end{align*}
Since the resulting graph is directed, we symmetrize: $L = \frac{1}{2}(L_0 + L_0^T)$.
Decomposing any $w \in \mathbb{R}^m$ as $w = \sum_{i=1}^m a_i u_i$ in the orthonormal eigenbasis of $L$, the Rayleigh quotient
\begin{align*}
    \frac{w^T L w}{w^T w} = \frac{\sum_{i=1}^m a_i^2 \mu_i}{\sum_{i=1}^m a_i^2}
\end{align*}
computes the mean frequency of $w$ as a graph signal.
We use the Rayleigh quotient as $f(g)$ for the vector $y = (y_1, \ldots, y_m)^T$.

\subsubsection{Fourier Transform of Data Projected onto Low-Dimensional Subspaces}
\label{sec:proj_fourier}

Following~\cite{xu-sb2}, we project the inputs $x_i$ onto vectors $u_j \in \mathbb{R}^d$ to obtain $a_{ji} = u_j^T x_i \in \mathbb{R}$, defining pseudo functions $h_j$ such that $h_j(a_{ji}) = y_i$.
These are ``pseudo'' functions since the mapping may not be well-defined (e.g., $a_{ji} = a_{jk}$ but $y_i \neq y_k$).
We compute their non-uniform discrete Fourier transforms and average over $z$ projection vectors:
\begin{align*}
    q(f) &= \frac{1}{z} \sum_{j=1}^z \mathcal{F}(h_j)(f), \qquad
    f(g) = \frac{\sum_{f \in F} |q(f)|^2 f}{\sum_{f \in F} |q(f)|^2}.
\end{align*}
We use the first $z = L$ left singular vectors of the input data matrix $X = [x_1 \cdots x_m] \in \mathbb{R}^{d \times m}$, whose span contains all columns of $X$ because the input functions are sums of $L$ trigonometric functions; see~\cref{app:data}.
However, the estimated spectrum depends on the choice of projection vectors: different orthonormal sets spanning the same space generally yield different frequency estimates, raising robustness concerns investigated in~\cref{sec:fourier_closing}.

\subsubsection{Comparison of Methods}
\label{sec:fourier_closing}

\begin{figure}[!ht]
    \centering
    \includegraphics[width=1.0\textwidth]{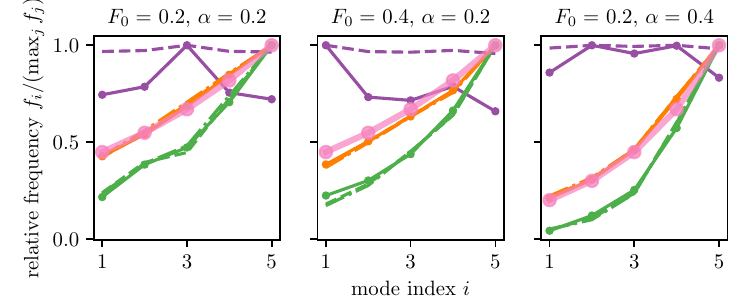}
    \caption{
    \textbf{Estimated frequency of synthetic data.}
    \textbf{Columns:} Different frequency families (synthetic datasets).
        \textbf{Colors and markers:}
        Pink: Dictated frequency $f_j = F_0 \exp(\alpha j)$,
        Orange: TV norm (dot-solid: $k=3$, dashed: $k=20$, dash-dotted: $k=50$),
        Green: LE (dot-solid: $k=3$, dashed: $k=50$),
        Purple: projection (left-singular vectors of $X$, dot-solid: $k=1$, dashed: $k=5$).
    }
    \label{fig:syntheticdata_frequency}
\end{figure}

\Cref{fig:syntheticdata_frequency} shows the estimated frequencies of the synthetic data using the three methods.
The TV norm accurately estimates the true relative frequency, and the LE is qualitatively correct.
The projection method fails to capture the trend of increasing frequencies.
We tested several projection vector sets (left-singular vectors of $X$, discrete sine basis), but none reliably captured the frequency trend.
We therefore use only the TV norm and LE in~\cref{sec:spectralbias}.

\end{document}